\documentclass{article}
\usepackage{ijcai25}

\usepackage{times}
\usepackage{soul}
\usepackage{url}
\usepackage[hidelinks]{hyperref}
\usepackage[utf8]{inputenc}
\usepackage[small]{caption}
\usepackage{graphicx}
\usepackage{amsmath}
\usepackage{amsthm}
\usepackage{booktabs}
\usepackage{algorithm}
\usepackage{algorithmic}
\usepackage[switch]{lineno}

\usepackage{appendix}
\usepackage{algorithm}
\usepackage{algorithmic}
\usepackage{booktabs}
\usepackage{multirow}
\usepackage{subcaption}

\title{Long-horizon Embodied Planning with Implicit Logical Inference and Hallucination Mitigation}

\author{
Siyuan Liu$^1$
\and
Jiawei Du$^1$\and
Sicheng Xiang$^{1*}$\and
Zibo Wang$^{1}$\footnote{These authors contributed equally.}\And
Dingsheng Luo$^1$\thanks{Corresponding author}\\
\affiliations
$^1$Peking University\\
\emails
billliu@pku.edu.cn, neodydu@stu.pku.edu.cn, 2152804@tongji.edu.cn,\\ 21371458@buaa.edu.cn, dsluo@pku.edu.cn
}

\begin{document}

\maketitle

\begin{abstract}
Long-horizon embodied planning underpins embodied AI. To accomplish long-horizon tasks, one of the most feasible ways is to decompose abstract instructions into a sequence of actionable steps. 
Foundation models still face logical errors and hallucinations in long-horizon planning, unless provided with highly relevant examples to the tasks. However, providing highly relevant examples for any random task is unpractical.
Therefore, we present ReLEP, a novel framework for \textbf{Re}al-time \textbf{L}ong-horizon \textbf{E}mbodied \textbf{P}lanning. ReLEP can complete a wide range of long-horizon tasks without in-context examples by learning implicit logical inference through fine-tuning. The fine-tuned large vision-language model formulates plans as sequences of skill functions. These functions are selected from a carefully designed skill library. ReLEP is also equipped with a Memory module for plan and status recall, and a Robot Configuration module for versatility across robot types. 
In addition, we propose a data generation pipeline to tackle dataset scarcity.
When constructing the dataset, we considered the implicit logical relationships, enabling the model to learn implicit logical relationships and dispel hallucinations.
Through comprehensive evaluations across various long-horizon tasks, ReLEP demonstrates high success rates and compliance to execution even on unseen tasks and outperforms state-of-the-art baseline methods.

\end{abstract}

\begin{figure*}[ht]
\centering
\includegraphics[width=0.8\textwidth]{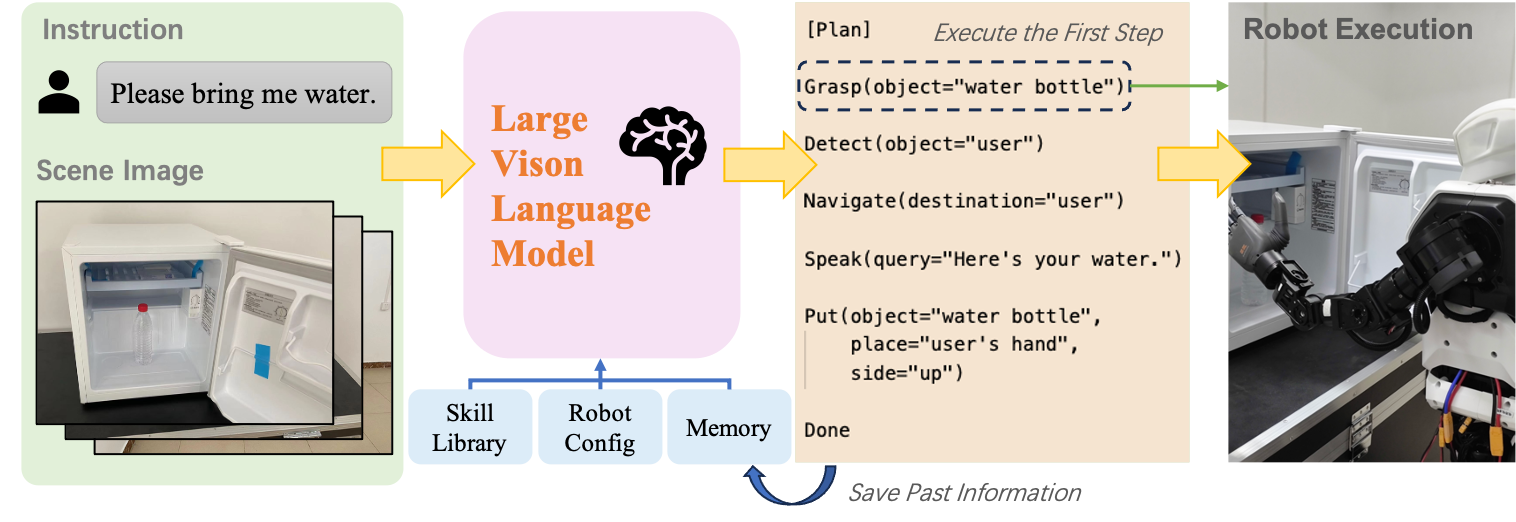} 
\caption{\textbf{An overview of ReLEP.} Given an instruction and a current scene image, a fine-tuned large vision-language model formulates plans as sequences of skill functions according to a skill library, a Memory module, and a Robot Configuration module. Then, the robot executes the first step and saves executed steps and past plans into the Memory module for subsequent rounds of planning.}
\label{fig_overview}
\end{figure*}

\section{Introduction}
The significant emergent abilities of foundation models to comprehend semantic information and reason make them good candidates for use as robot brains. However, even with the help of foundation models, embodied tasks are still difficult to carry out. 

Foundation models use language as a medium to interact with the environment. It is hard to ground tasks in natural language into robot action space. There are generally three ways to achieve the grounding of embodied tasks. 
The first is to fine-tune the foundation model to directly output control signals in text form. This method can only be used in simple control like primitive autonomous driving \cite{xu2023drivegpt4}, and the performance is not guaranteed. 
The second approach is vision-language-action models \cite{brohan2023rt,belkhale2024rt}. It assigns action controls to rarely used tokens or integer tokens in the original foundation models and names them action tokens, then trains these models to predict action tokens given language input. 
The third way is more cost-efficient. It uses pre-defined skills as APIs, and let the foundation models predict a sequence of skills, then execute these skills in succession \cite{ahn2022can,huang2023instruct2act}. 

Another difficulty lies in task planning. Tasks such as ``bring me a bottle of water" can be difficult to carry out in an end-to-end manner, as they often require multiple steps and contextual understanding. These long-horizon tasks usually need to be broken down into multiple short-term low-level tasks in the first place. 
Previous works show that advanced foundation models are already capable of decomposing simple long-chain tasks \cite{hu2023look,zhi2024closed}. 
However, they can currently only be applied to a small skill set, which limits them to completing a narrow array of tasks. When given more complex skills, the models cannot fully understand the logical relationships between skills through limited examples, leading to logical errors in planning and hallucinations of using self-fabricated skills.
They require highly relevant task-specific in-context examples to generate tractable plans for later execution (see Section \ref{sub_c_l}). 
The limited comprehension of foundation models regarding the skills and their fluid expressions is the underlying cause of the aforementioned issues.

Since creating highly relevant task-specific examples is impractical, we seek to address the problem through model fine-tuning. To this end, we require a dataset that encompasses the implicit logical relationships between skills.
Most embodied planning methods are carried out in simulated environments \cite{shridhar2020alfred,huang2023instruct2act,skreta2024replan} due to the lack of real-world data. EgoCOT \cite{mu2024embodiedgpt} is one of the few real-world datasets for embodied planning, but the skills in the dataset are free-form and difficult to reuse.

In this work, we introduce ReLEP, a universal framework for real-time long-horizon embodied planning via large vision-language models. An overview of ReLEP is shown in Figure \ref{fig_overview}.
The brain of ReLEP is a fine-tuned large vision-language model which is aligned to decompose long-horizon tasks into a sequence of pre-defined skill functions. 

We present a pipeline to construct our dataset with real-world images using GPT-4V \cite{gpt-4v}. In constructing the dataset, the logical relationships between the skills are considered, such as the logical sequence of skills, skill parameters, robot embodiment, and hallucinations.
In this way, the model can leverage the learned implicit logical inference to better understand the logical relationships between skills and their usage, while reducing hallucinations.

ReLEP has a carefully constructed skill library, including nine interactive and non-interactive skills. By combining these skills, ReLEP can accomplish dozens of long-horizon tasks, such as manipulation tasks, pick-and-place tasks, interaction tasks, navigation tasks, inspection tasks, delivery tasks, and more.

To maintain consistency in multi-step planning, ReLEP introduces a Memory Module to store past information and the robot's state. A Robot Configuration Module is used to prompt the model to account for the logical issues introduced by the robot's embodiment in planning.
With the help of these modules, ReLEP can be applied to robots with different embodiments and skill libraries. 

Comprehensive evaluations have been conducted to test the performance of foundation models and ReLEP.
We first conducted an in-context experiment to demonstrate the reliance of foundation models on highly relevant examples in long-horizon planning tasks. 
Then, we compared ReLEP with the state-of-the-art method VILA \cite{hu2023look} and model GPT-4V and as baselines.
In the simulation environment AI2-THOR \cite{kolve2017ai2}, real-time comparative experiments were carried out on ten long-horizon everyday tasks, including both tasks within and outside of the training set.
In addition, we collected images from the real world as input to test whether ReLEP also exhibits strong planning capabilities in real-world scenarios. 
To validate the effectiveness of the key modules, we also performed ablation studies.
The results indicate that ReLEP outperforms all baseline methods in different ways and demonstrate the effectiveness of the proposed framework.

The main contributions of our work are as follows:

\begin{itemize}
\item We discovered that foundation models using in-context learning methods encounter logical errors and hallucinations in long-horizon embodied planning. We point out that fine-tuning the model with implicit logical relationships can effectively solve these issues. Additionally, we fine-tuned a large vision-language model with implicit logical inference and hallucination mitigation.

\item A real-time long-horizon embodied planning framework named ReLEP, which is able to complete a wide range of long-horizon embodied tasks with implicit logical inference and hallucination mitigation. The framework can be adopted for robots with different embodiments and skill libraries.

\item A pipeline to generate real-world embodied planning dataset with the help of GPT-4V, and a 5K + 24K real-world embodied planning dataset constructed with implicit logical relationships. This fills a gap in real-world data for long-horizon embodied tasks and is of great significance for the development of embodied agents in real-world environments.

\end{itemize}

\section{Related Work}
\subsection{Large Vision-Language Models}
\label{RW_LVM}
The outstanding capabilities of large language models inspire researchers to build large multimodal models for multimodal instruction-following tasks. Large vision language models such as BLIP-2 \cite{li2023blip}, LLaVA \cite{liu2024visual},  InstructBLIP \cite{dai2023instructblipgeneralpurposevisionlanguagemodels}, MiniGPT-v2 \cite{chen2023minigpt}, Qwen-VL \cite{bai2023qwen}, LLaVA-1.5\cite{liu2024improved} and GPT-4V \cite{gpt-4v} have shown their capabilities in classification, detection, segmentation, captioning, and visual question answering. One of the most exciting discoveries is the possibility that large vision language models can perform embodied tasks like embodied question answering and embodied planning. 

\subsection{Long-horizon Embodied Planning via Large Vision-Language Models}
TaPA \cite{wu2023embodied} collects multiple RGB images in simulator scenes from different viewpoints, and utilizes detectors to detect a list of objects. This object list together with task instructions in natural language is passed into a fine-tuned large language model to generate a sequence of task plans. TaPA can solve long-horizon tasks that require robot navigation.

PaLM-E \cite{driess2023palm} combines vision, continuous state estimation, and natural language instructions as input to generate a text plan, using a classic ViT \cite{dosovitskiy2020image,dehghani2023scaling} and a large language model framework. The model itself cannot execute low-level control. Furthermore, it lacks the ability to process intricate instructions.

EmbodiedGPT \cite{mu2024embodiedgpt} introduces a new model architecture embodied-former as a bottleneck to pass the most relevant visual information to the language model. EmbodiedGPT is trained with embodied plannings generated by ChatGPT based on videos from the Ego4D dataset \cite{grauman2022ego4d}. It can generate embodied plans in natural language with an input image, or explain what embodied plan is performed by a robot in a given video. 

With abundant training and data resources, the Tech Giants have launched powerful large vision language models such as GPT-4V. These models can achieve successful planning for simple tasks without further training. With appropriate prompting and few-shot examples, they are able to complete real-world embodied tasks planning.

VILA \cite{hu2023look} leverages GPT-4V to generate an executable step-by-step plan, and performs closed-loop embodied tasks with a fixed robot arm. VILA can perform reasoning related to spatial layout and object attributes on visual input.

More recently, COME-robot \cite{zhi2024closed} utilizes GPT-4V for open-ended reasoning and re-planning in real-world environments. Instead of generating plans in text, it directly queries GPT-4V for predefined plan code, as well as the reasoning process in natural language from a first-person perspective. Experiments are carried out with a mobile robot arm.

These works use GPT-4V as brain to generate plans in predefined form, then execute them with low-level action APIs. These actions are often limited and simple. For example, COME-robot has only six action functions, including navigate, grasp, place, and functions for exploration. Therefore, the types of tasks they can complete are limited. 
This framework works fine for a small library of action functions. However, when the library scales up, GPT-4V may find it difficult to master the usage of these functions and fail to generate appropriate plans without highly relevant task-specific in-context examples.

\section{Methodology}
We first introduce the formulation of the real-time embodied planning problem in Section \ref{sPF}. The details of our framework are discussed in Section \ref{sF}. Finally, we present the pipeline we used to generate real-world embodied task planning dataset in Section \ref{sDA}.

\subsection{Problem Formulation}
\label{sPF}
\subsubsection{Long-horizon Embodied Planning}
Given an embodied task $E$ in natural language and an image of the scene $I$, the agent $M$ decomposes $E$ into a sequence of skill functions $P$. These skill functions belong to a set of pre-defined skill library $\Pi=\{s_{1},s_{2},...,s_{n}\}$. By executing $P$ in succession, the agent should be able to complete task $E$.
\begin{equation}
    P = M(E, I, \Pi)
\end{equation}

\subsubsection{Real-time Embodied Planning}
At time $t_{0}$, the agent is given an embodied task $E$ in natural language and an image of the current scene $I_{0}$. Then the agent $M$ decomposes $E$ into a sequence of skill functions $P_{0}$. At this point, the first round of embodied task planning is complete.
After the robot executes the first skill function of $P_{0}$ by generating a trajectory $\tau$ to control the robot or performing a non-interactive function, the environment changes and another image of the scene $I_{1}$ at time $t_{1}$ is given. The agent then decomposes $E$ into another sequence $P_{1}$, referring to $P_{0}$.

\begin{equation}
    P_{i} = M(E, I_{i}, \Pi, h_{i-1})
\end{equation}
where $P_{i}$ is the plan generated by the agent and $I_{i}$ is the image of the scene at time $t_{i}$. $h_{i-1}$ denotes the past information accumulated up to time $t_{i}$.

\subsection{ReLEP}
\label{sF}
The real-time long-horizon embodied planning framework consists of five parts: a large vision-language model for planning, three storage modules for prompting, and an execution module for interacting. 

At time $t_{0}$ an embodied task $E$ and the initial image $I_{0}$ are passed into the framework. The large vision language model generates its initial plan $P_{0}$ with guidance in the Skill Library and the Robot Configuration. Then, the execution module performs the first step of the plan by calling the corresponding skill function. Finally, $P_{0}$ is recorded in the Memory module as well as the finished steps and robot status.

For the upcoming time stamp $t_{i}$, the same task $E$ and the current image $I_{i}$ are passed into the framework. This time, the model refers to all three modules to make $P_{i}$. Then the first step of $P_{i}$ is executed and the information is recorded as in the first round.

The loop will break at time $t_{i+1}$ if $P_{i}$ only contains a \textbf{Done} state.

The pseudo code of the entire process is shown in Algorithm \ref{algorithm_r_p} for better understanding.

\begin{algorithm}[tb]
\caption{ReLEP}
\label{algorithm_r_p}
\textbf{Require}: Initial visual image $I_{0}$, an embodied task instruction $E$ and a skill library $\Pi$
\begin{algorithmic}[1] 
\STATE Let $t=0$, $P_{0}=$[], $F=$[]
\STATE $P_{0}=$LVLM$((E,\Pi), I_{0})$
\WHILE{$P_{t}[0]\neq$"Done"}
\STATE Execute skill $\Pi(P_{t}[0])$
\STATE Add $P_{t}[0]$ to $F$
\STATE $t=t+1$
\STATE $I_{t}=$GetCurrentImage()
\STATE $P_{t}=$LVLM$((E,\Pi,P_{t-1},F), I_{t})$
\ENDWHILE
\end{algorithmic}
\end{algorithm}

\subsubsection{Large Vision-Language Model as Brain}
The center of the framework is a large vision-language model that predicts plans based on information from other modules. 
Compared to large language models in embodied tasks, large vision language models take visual input directly instead of converting it into texts and then passing it into large language models \cite{huang2022language,song2023llm,liang2023code,wu2024mldt}. This helps the model see the scene itself, avoiding information loss and misunderstanding \cite{hu2023look}.

We choose to fine-tune the instruction-following model LLaVA-1.6-7B \cite{liu2024llavanext}. 
We fine-tuned the model from a checkpoint with LoRA \cite{hu2021lora}. We first trained the model for one epoch using our 5K first-round planning data. 
In order to acquire the subsequent-round planning ability, we continued to train the model for two epochs using 5K first-round + 5K subsequent-round data and 5K first-round + 10K subsequent-round data respectively. The subsequent-round data for each epoch were randomly selected from the full 24K subsequent-round data.

By training with data containing implicit logical relationships, the model can acquire implicit logical inference capabilities and effectively reduce hallucinations.

\subsubsection{Skill Library}
The skill library is one of the most important parts of our framework. It functions as the limbs of the embodied agent to interact with the environment. 

We have carefully designed nine skills that can be combined to complete a wide range of tasks in daily life.
Details of these skills are listed in Appendix.

Furthermore, we designed two states at the end of each plan: \texttt{Done} and \texttt{Pending}. \texttt{Done} state indicates that the task is completed, while \texttt{Pending} means that the task is not completed and the planning can only continue after the status of a certain step is confirmed. 
If a plan ends with \texttt{Done}, it means that the task can be completed according to the information collected so far. Otherwise, it will end with \texttt{Pending}.

\subsubsection{Memory}
If the agent does not consider previous plans, the plans made at different time will show inconsistency. Therefore, the agent should refer to previous plans in its Memory in subsequent rounds of planning.
More importantly, we found that it is difficult for the agent to determine whether a step is finished according to the image (see Section \ref{sub_a_s}). Thus, when a step is executed, we add it to the Finished Steps in Memory module to ask the agent to proceed to the next step. 
In addition, the agent may forget its status during a long-chain mission. Inferring the implicit status from the plan is not easy for the agent. To this end, the Memory module should contain the status of the robot.

\subsubsection{Robot Configuration}
Robots with different embodiments may not be able to share the same skill library. For instance, a quadruped robot dog cannot perform \texttt{Grasp} since it has no robot arm. In terms of planning, a single-armed robot cannot perform two \texttt{Grasp}s in a row, compared to a humanoid robot.

With this in mind, we introduce the Robot Configuration module, which uses natural language to describe the robot configuration, prompting the agent to refer to this module while planning. 

\subsection{Dataset Acquisition}
\label{sDA}
The dataset we used for fine-tuning consists of two parts: 5K first-round planning data with past information and 24K subsequent-round data without. 

The pipeline to generate the embodied planning dataset shown in Appendix takes two steps: using images to generate tasks, then using the paired image and task to generate the plan. 
In the end, we will obtain a set of triplets (image, task, plan) then re-form them into dialogues of the first round and subsequent rounds.

\subsubsection{Task Generation}
We first collected a set of indoor object detection datasets \cite{indoorgit,indoorrobo} from the Internet. Object detection has a large foundation of image data, and images from the indoor object detection dataset are often more closely related to embodied tasks.
Then we manually filtered out the images whose scenes are not suitable for generating embodied tasks.
After obtaining the filtered images, we queried GPT-4V to generate five tasks that a robot may perform on the scene of those images.
Sometimes, GPT-4V may hallucinate and generate tasks related to objects that are not in the scene.
We then modified or deleted these tasks on the basis of hallucination and feasibility.

\subsubsection{Plan Generation}
\label{subsub_p_g}
With the paired images and tasks that we generated from the first step, we constructed a simplified prompt for embodied planning to query GPT-4V to create a plan in a specified format. 
However, GPT-4V sometimes makes logical errors or it may hallucinate and use undefined skills.
Finally, we manually modified unreasonable plans according to the implicit logical relationship between each skill, as well as the physical constraints of the robot embodiment.

\begin{table*}[ht]
    \begin{center}
    \begin{tabular}{c|c c c c c c c c c}
        \toprule
        \addlinespace
        \multirow{2}{*}[-0.3cm]{\textbf{Task}} & \multicolumn{3}{c}{GPT-4V zero-shot} & \multicolumn{3}{c}{GPT-4V loosely relevant} & \multicolumn{3}{c}{GPT-4V highly relevant} \\ 
        \addlinespace
        \cmidrule(r){2-4} \cmidrule(r){5-7} \cmidrule(r){8-10}
        \addlinespace
         & PSR & ESR & LC & PSR & ESR & LC & PSR & ESR & LC \\
        \addlinespace
        \midrule
        Turn on Lights & \textbf{10/10} & 0/10 & 0.0 & \textbf{10/10} & \textbf{10/10} & \textbf{100.0} & \textbf{10/10} & \textbf{10/10} & \textbf{100.0}\\
        \addlinespace
        Push Chair & 8/10 & 0/10 & 0.0 & 8/10 & 0/10 & \textbf{100.0} & \textbf{10/10} & \textbf{10/10} & \textbf{100.0} \\
        \addlinespace
        Bring Water & 8/10 & 0/10 & 0.0 & 9/10 & 0/10 & \textbf{100.0} & \textbf{10/10} & \textbf{10/10} & \textbf{100.0} \\
        \addlinespace
        Bring Book & 8/10 & 0/10 & 0.0 & \textbf{10/10} & 0/10 & \textbf{100.0} & \textbf{10/10} & \textbf{10/10} & \textbf{100.0} \\
        \addlinespace
        Throw Trash & 0/10 & 0/10 & 0.0 & 0/10 & 0/10 & 95.4 & \textbf{10/10} & \textbf{10/10} & \textbf{100.0} \\
        \addlinespace
        Fill Kettle & 0/10 & 0/10 & 0.0 & 6/10 & 0/10 & 99.3 & \textbf{10/10} & \textbf{10/10} & \textbf{100.0} \\
        \addlinespace
        Put Mug & 3/10 & 0/10 & 0.0 & 9/10 & 0/10 & 84.9 & \textbf{10/10} & \textbf{8/10} & \textbf{100.0} \\
        \midrule
        \addlinespace
        \textbf{Total} & 52.9 & 0.0 & 0.0 & 74.3 & 14.3 & 96.0 & \textbf{100.0} & \textbf{97.1} & \textbf{100.0} \\
        \addlinespace
        \bottomrule
    \end{tabular}
    \caption{\textbf{Evaluation results of GPT-4V with in-context examples of varying relevance of long-horizon embodied planning.} GPT-4V zero-shot predicts freeform plans with zero LC. GPT-4V with loosely relevant examples has improved LC but makes logical errors. GPT-4V with highly relevant examples can make logical plans with high success rate. All decimal values are presented as percentages.}
    \label{table:incontext}
    \end{center}
\end{table*}

\section{Experiments}
We first tested GPT-4V with in-context examples of varying relevance in long-horizon embodied planing in Section \ref{sub_c_l}.
Then we conducted comparative experiments in a simulated environment and with collected real-world images in Section \ref{sub_c_e}.
Finally, we conducted ablation experiments on ReLEP's model and its Memory module in Section \ref{sub_a_s}.
\begin{figure}[ht]
	\centering
	\begin{subfigure}[b]{0.49\linewidth}
		\centering
		\includegraphics[width=0.9\linewidth]{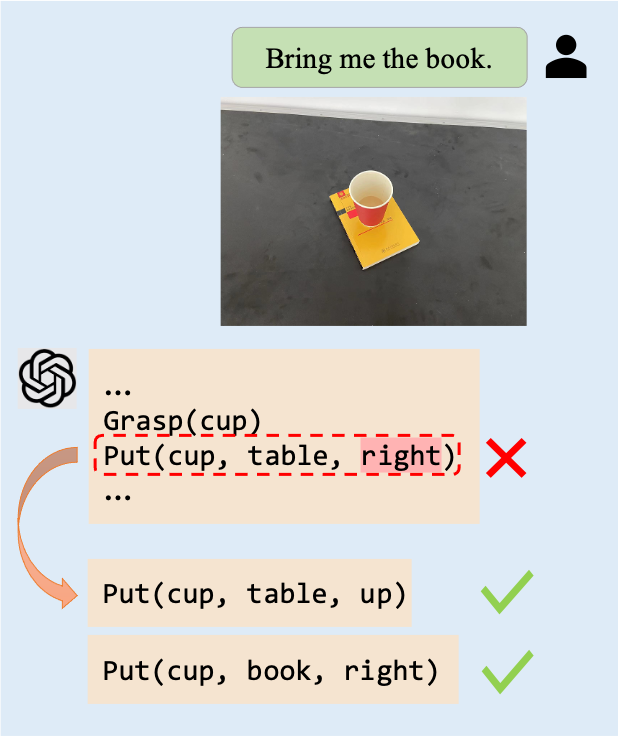}
        \subcaption{Logical understanding errors}
	\end{subfigure}
	\begin{subfigure}[b]{0.49\linewidth}
		\centering

		\includegraphics[width=0.9\linewidth]{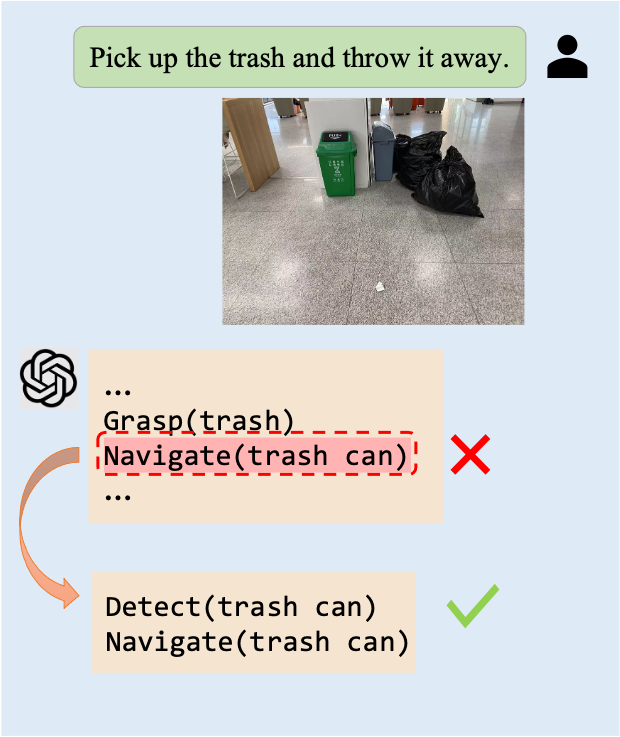}
        \subcaption{Skill combination errors}
	\end{subfigure}
	\begin{subfigure}[b]{0.49\linewidth}
		\centering
		\includegraphics[width=0.9\linewidth]{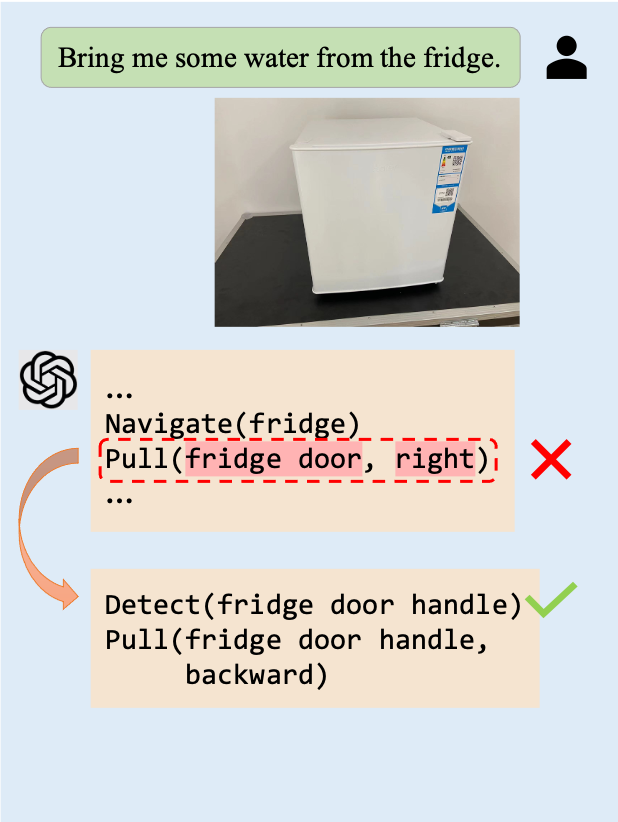}
        \subcaption{Logical planning errors}
	\end{subfigure}
	\begin{subfigure}[b]{0.49\linewidth}
		\centering
		\includegraphics[width=0.9\linewidth]{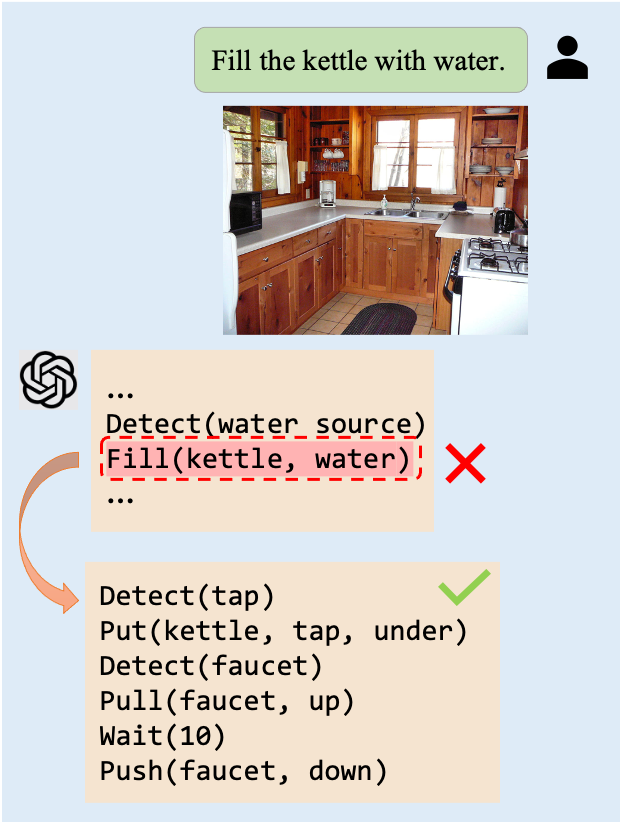}
        \subcaption{Hallucinations}
	\end{subfigure}
    \caption{\textbf{Four types of errors GPT-4V made in long-horizon embodied planning with loosely relevant examples.} Logical understanding errors are misinterpretations of individual skills. Missing skills can lead to skill combination errors. Logical mistakes result in logical planning errors. Hallucinations include fabricating undefined skills.}
    \label{fig_cl}
\end{figure}
\subsection{Can Foundation Models Master the Usage of Skills?}
\label{sub_c_l}
To test how well foundation models master the use of skills and logical relationships through in-context learning, we provided GPT-4V with in-context examples of varying relevance to the tasks and conducted comparative experiments. That is, highly relevant, loosely relevant, and zero-shot.
Each example includes a different task, a corresponding scene image, and a correct plan.

Highly relevant examples are carefully constructed according to the given tasks. Some of them are constructed by replacing the subject of the task, so the structures of the plans are mostly the same.
Weakly relevant examples for all tasks are the same, which is a simple Bring Water task presented in Appendix.

We introduce \textit{Plausible Success Rate (PSR)}, \textit{Executable Success Rate (ESR)}, and \textit{Language Compliance (LC)} to measure the performance of each method. 
Since the output skill sequences may not meet the strict requirements for execution, we count them as plausibly successful as long as they are reasonable regardless of the execution.
\textit{Language Compliance} was designed by \cite{li2024muep} to measure the structuredness and compliance of the skill generated by the foundation models. It is given as $LC=\hat{s}/s^{*}$, where $\hat{s}$ is the number of valid skills and $s^{*}$ is the total number of skills.

We tested the three methods on seven long-horizon embodied planning tasks, and the results are shown in Table \ref{table:incontext}.

Without in-context examples demonstrating the format of the output and the logical relationships between the skills, GPT-4V can make plausible plans but with errors in detail. 
GPT-4V with highly relevant, manually crafted examples performs well in all tasks.

GPT-4V with loosely related in-context examples can sometimes produce textually plausible plans, but due to logical errors and hallucinations, they are not executable. 
We categorize these errors into four types in Figure \ref{fig_cl}: logical understanding errors, skill combination errors, logical planning errors, and hallucinations. 

\texttt{Put(cup, table, right)} puts the cup on the right side of the table, on the ground. While the model assumes it should be placed on the right side of the tabletop. This results in a logical understanding error.
Before navigating to a destination, the agent should first detect it. Not detecting will result in failure of the next step, which is a typical skill combination error.
Pulling the fridge door instead of the fridge door handle and pulling in the wrong directions are logical planning errors.
Not knowing how to combine the skills provided to fill the kettle with water, the model hallucinated and fabricated an undefined skill \texttt{Fill}.

All of these errors and hallucinations are caused by the inability of GPT-4V to understand the logical relationships between skills without examples that are highly relevant to the task. However, building such task-specific examples for each task is no different from manually planning.
 
\begin{table*}[ht]
    \begin{center}
    \begin{tabular}{c|c c c c c c c c c c c c}
        \toprule
        \addlinespace
        \multirow{2}{*}[-0.3cm]{\textbf{Task}} & \multicolumn{3}{c}{VILA$^{*}$} & \multicolumn{3}{c}{GPT-4V$^{*}$} & \multicolumn{3}{c}{ReLEP+GPT-4V} & \multicolumn{3}{c}{ReLEP}\\ 
        \addlinespace
        \cmidrule(r){2-4} \cmidrule(r){5-7} \cmidrule(r){8-10} \cmidrule(r){11-13}
        \addlinespace
         & IPSR & SSR & SR & IPSR & SSR & SR & IPSR & SSR & SR & IPSR & SSR & SR \\
        \addlinespace
        \midrule
        Close Laptop & 7/10 & 30/41 & 7/10 
        & \textbf{10/10} & 29/38 & 1/10
        & \textbf{10/10} & 32/38 & 2/10
        & \textbf{10/10} & \textbf{50/50} & \textbf{10/10} \\
        \addlinespace
        Push Chair & 6/10 & 12/21 & 1/10 
        & \textbf{10/10} & 20/30 & 0/10
        & \textbf{10/10} & 20/27 & 3/10 
        & \textbf{10/10} & \textbf{40/40} & \textbf{10/10} \\
        \addlinespace
        Turn on Lights & 9/10 & 25/18 & 7/10 
        & \textbf{10/10} & 25/30 & 50.0
        & \textbf{10/10} & 26/30 & 70.0
        & \textbf{10/10} & \textbf{30/30} & \textbf{10/10} \\
        \addlinespace
        Look into Mirror & 6/10 & 23/27 & 8/10 
        & \textbf{10/10} & 28/32 & 6/10
        & \textbf{10/10} & 22/27 & 5/10
        & \textbf{10/10} & \textbf{30/30} & \textbf{10/10} \\
        \addlinespace
        Put Mug & 0/10 & 42/52 & 0/10
        & \textbf{10/10} & 61/71 & 0/10
        & \textbf{10/10} & 45/56 & 0/10
        & \textbf{10/10} & \textbf{80/80} & \textbf{10/10} \\
        \midrule
        \addlinespace
        Take Egg & 4/10 & 14/24 & 0/10
        & \textbf{10/10} & 72/81 & 5/10
        & \textbf{10/10} & 36/44 & 2/10
        & \textbf{10/10} & \textbf{110/110} & \textbf{10/10} \\
        \addlinespace
        Fill Mug & 0/10 & 32/42 & 0/10
        & \textbf{10/10} & 70/80 & 0/10
        & \textbf{10/10} & 44/54 & 0/10
        & \textbf{10/10} & \textbf{100/100} & \textbf{10/10} \\
        \addlinespace
        Flush Knife & 1/10 & 10/20 & 0/10
        & \textbf{10/10} & 70/78 & 2/10
        & 9/10 & 67/76 & 1/10
        & \textbf{10/10} & \textbf{91/92} & \textbf{9/10} \\
        \addlinespace
        Pull Curtain & 5/10 & 11/19 & 2/10
        & \textbf{10/10} & 25/28 & 7/10
        & \textbf{10/10} & 23/28 & 6/10
        & \textbf{10/10} & \textbf{40/40} & \textbf{10/10} \\
        \addlinespace
        Lift Toilet Seat & 6/10 & 14/23 & 1/10
        & \textbf{10/10} & 31/38 & 3/10
        & \textbf{10/10} & 33/37 & 6/10
        & \textbf{10/10} & \textbf{40/40} & \textbf{10/10} \\
        \midrule
        \addlinespace
        \textbf{Total (\%)} & 44.0 & 71.7 & 26.0 
        & \textbf{100.0} & 85.2 & 29.0 
        & 99.0 & 83.5 & 32.0
        & \textbf{100.0} & \textbf{99.8} & \textbf{99.0}\\
        \addlinespace
        \bottomrule
    \end{tabular}
    \caption{\textbf{Quantitative evaluation results on real-time long-horizon tasks in simulated environment AI2-THOR.} ReLEP demonstrates superior performance both in seen tasks (top half) and unseen tasks (bottom half). The total steps of SSR of each task and method depends on which step the failure occurs during each run. Early failure can lead to fewer steps of SSR.}
    \label{table:sim}
    \end{center}
\end{table*}
\subsection{Comparative Experiments}
\label{sub_c_e}
\subsubsection{Baselines and Metrics}
We choose GPT-4V$^{*}$, ReLEP+GPT-4V, and the VILA \cite{hu2023look} method as our baselines.

We equip VILA with the same skill library as ReLEP, and utilize its original framework for planning, referred as VILA$^{*}$.

GPT-4V$^{*}$ uses the ReLEP framework, but replaces the fine-tuned model with GPT-4V for all rounds of planning. In the first round of planning, GPT-4V$^{*}$ is provided with highly relevant, manually crafted in-context examples. For in the previous experiment, GPT-4V was unable to make executable plans when only loosely relevant examples were provided. It should be emphasized that this is a significant enhancement to GPT-4V$^{*}$, and poses an even greater challenge for ReLEP.

Apart from using GPT-4V for subsequent rounds of planning, the rest of ReLEP+GPT-4V is identical to ReLEP. 

We compared the success rate of the initial plans (\textit{Initial Plan Success Rate, IPSR}) made on the designed tasks, demonstrating the ability of first-round planning. 
In addition, to verify the ability of subsequent-round planning, we also display the success rate of each planning step as (\textit{Step-wise Success Rate, SSR}).
Finally, we report the success rate of the entire process (\textit{Success Rate, SR}).

\subsubsection{Real-time Experiment in Simulated Environment}
We selected six scenes in the AI2-THOR simulation environment and designed ten long-horizon embodied tasks for real-time comparative experiments.
These tasks include both tasks from the training set and tasks outside the training set. Tasks outside the training set include unseen objects and unseen tasks.
Furthermore, all the scenes used in the experiment are unseen.

We present the quantitative evaluation results in Table \ref{table:sim}.

VILA$^{*}$ tends to overthink and discusses a variety of scenarios, resulting in longer inference time. Moreover, most of the output is in natural language, with the actual plan comprising only a small portion. VILA$^{*}$ often ignores the navigation skill and manipulates objects without approaching.

With the help of highly relevant examples, GPT-4V$^{*}$ performs well in the first round of planning. However, GPT-4V sometimes gets stuck at \texttt{Detect} or \texttt{Navigation} in subsequent rounds of planning and repeatedly executes the same skill. 
This may be because it is difficult for the model to confirm whether these steps have been completed based on the image information. 
In fact, the prompts already include all the finished steps, but the model does not refer to them.

ReLEP has similar IPSR compared to GPT-4V with highly relevant examples, demonstrating the effectiveness of fine-tuning with the method incorporating implicit logical relationships. 

ReLEP achieves high success rates on all tasks, suggesting the implicit logical inference and hallucination mitigation capabilities of our approach. 
Even in unseen tasks, ReLEP maintains a high success rate, indicating the robustness of the method.
Its higher SSR and SR than other baselines showcase ReLEP's outstanding ability in real-time long-horizon planning and its mastery of skill sets.

\begin{table}[tb]
    \begin{center}
    \begin{tabular}{l| c c c}
        \toprule
        \addlinespace
        Methods & IPSR & SSR & SR \\
        \addlinespace
        \midrule
        VILA$^{*}$ & 61.4 & 81.1 & 58.6  \\
        \addlinespace
        GPT-4V$^{*}$ & 82.9 & 77.5 & 22.9 \\
        \addlinespace
        ReLEP+GPT-4V & \textbf{94.3} & 83.6 & 34.3 \\
        \addlinespace
        ReLEP & \textbf{94.3} & \textbf{97.7} & \textbf{94.3} \\
        \addlinespace
        \bottomrule
    \end{tabular}
    \caption{\textbf{Evaluation results with real-world images on long-horizon tasks.} ReLEP still holds a higher success rate, highlighting its effectiveness in real-world environments. All values are presented as percentages.}
    \label{table:realworld}
    \end{center}
\end{table}

\subsubsection{Real-world Experiment Using Collected Images}
We designed eight long-horizon embodied tasks in real-world environment to verify ReLEP's ability in real-world planning,
In this experiment, we did not execute the steps in the plans. Instead, we collected scene images after each executed step for offline testing and manually assessed the success rate.
That is to say, the captured images are what the scene should look like after the steps are executed.
Because the inputs of the models and the evaluation criteria are essentially the same, this experiment can also verify the effectiveness of the methods.
The evaluation results are shown in Table \ref{table:realworld}.

ReLEP still demonstrates a higher success rate compared to other baselines, highlighting its effectiveness in real-world environments.

\begin{table}[tb]
    \begin{center}
    \begin{tabular}{l| c c c c}
        \toprule
        \addlinespace
        Methods & LC & IPSR & SSR & SR \\
        \addlinespace
        \midrule
        ReLEP (ours) & \textbf{100.0} & \textbf{100.0} & \textbf{99.8} & \textbf{99.0} \\
        \addlinespace
        w/o Memory & \textbf{100.0} & \textbf{100.0} & 47.8 & 0.0 \\
        \addlinespace
        w/o FT (zero-shot) & 0.0 & 0.0 & 0.0 & 0.0 \\
        \addlinespace
        w/o FT (loosely relevant) & 48.6 & 4.0 & 46.2 & 0.0 \\
        \addlinespace
        \bottomrule
    \end{tabular}
    \caption{\textbf{Ablations over the large vision-language model and the Memory module.} Without the Memory module, ReLEP fails on subsequent rounds of planning and has a lower SSR. Without fine-tuning, the model makes logical errors even with loosely relevant examples and has low LC. All values are presented as percentages.}
    \label{table:ablation}
    \end{center}
\end{table}

\subsection{Ablation Study}
\label{sub_a_s}
In the ablation study, we ablate the large vision-language model to demonstrate the effectiveness of fine-tuning, and the Memory module to prove its necessity in subsequent-round planning. Results are presented in Table \ref{table:ablation}.

To ablate the large vision-language model, we directly use LLaVA-1.6-7B to predict the plans.
We compared ReLEP with both LLaVA-1.6-7B zero-shot and LLaVA-1.6-7B with loosely relevant examples.
Given the same prompt as ReLEP, LLaVA-1.6-7B zero-shot generates plans in freeform like GPT-4V zero-shot. This was expected since the model cannot possibly predict the required format without examples.
With loosely relevant examples, LLaVA-1.6-7B performs worse than GPT-4V, with more logical errors and hallucinations. Moreover, it fails on subsequent rounds of planning with low Language Compliance.
ReLEP's higher SR and LC suggest that fine-tuning the large-vision language model is necessary. It helps the model learn implicit logical inference and better understand how each module works.

To verify the effectiveness of Memory module, we compared ReLEP with a version of ReLEP without Memory module on the long-horizon embodied tasks in the simulated environments.
ReLEP without Memory module fails to predict new plans after steps like \texttt{Detect} and \texttt{Navigate} because it is difficult for the model to determine whether such steps are finished solely according to the image. 
In other words, some skills do not change the scene image or cause minimal changes in the scene, resulting in the same or similar inputs between steps. Since the framework does not record the completed steps, the model cannot distinguish between the two states.
Therefore, it stuck in the loop to repeatedly detect an object like GPT-4V$^{*}$ did. 

Nevertheless, we proceeded our experiments by providing subsequent images to verify if it could break the loop based on obvious changes in the image.
As a result, obvious changes like the refrigerator door being opened can be detected by ReLEP without the Memory module, but not for small changes like getting closer to the target.

In conclusion, the above facts and ReLEP's higher SSR demonstrates the effectiveness of the Memory Module.

\section{Conclusion}
We pointed out that foundation models using in-context learning face logical errors and hallucinations in long-horizon embodied tasks.
To this end, we proposed to fine-tune a large visual-language model with implicit logical relations.
Furthermore, we proposed a novel framework for real-time long-horizon embodied planning called ReLEP. It plans with a large vision-language model according to the environment image. 
We also proposed a pipeline for real-world embodied planning data generation and constructed a real-world long-horizon embodied planning dataset with implicit logical relationships.
Comprehensive experiments are conducted and the results demonstrate the effectiveness of ReLEP and suggest that it outperforms the SOTA method and the GPT-4V model.

However, the real-world application of ReLEP requires robust and universal algorithms as well as low-level control policies. 
Although obtaining these algorithms remains a challenge, current advancements in general tasks such as object detection, grasping, and navigation \cite{liu2025grounding,wan2023unidexgrasp++,chang2023goat} have created promising prospects for the implementation of skill functions.
In addition, we only constructed a small real-world embodied planning dataset with 5K basic data due to limited resources. Hopefully, more researchers will be joining this cause and build larger datasets with better quality.

\bibliographystyle{named}
\bibliography{ijcai25}

\newpage
\appendix

\onecolumn
\renewcommand\thetable{\Alph{section}\arabic{table}}
\renewcommand\thefigure{\Alph{section}\arabic{figure}} 
\setcounter{table}{0}
\setcounter{figure}{0}

\section{Appendix}
\begin{figure*}[ht]
\centering
\includegraphics[width=0.8\textwidth]{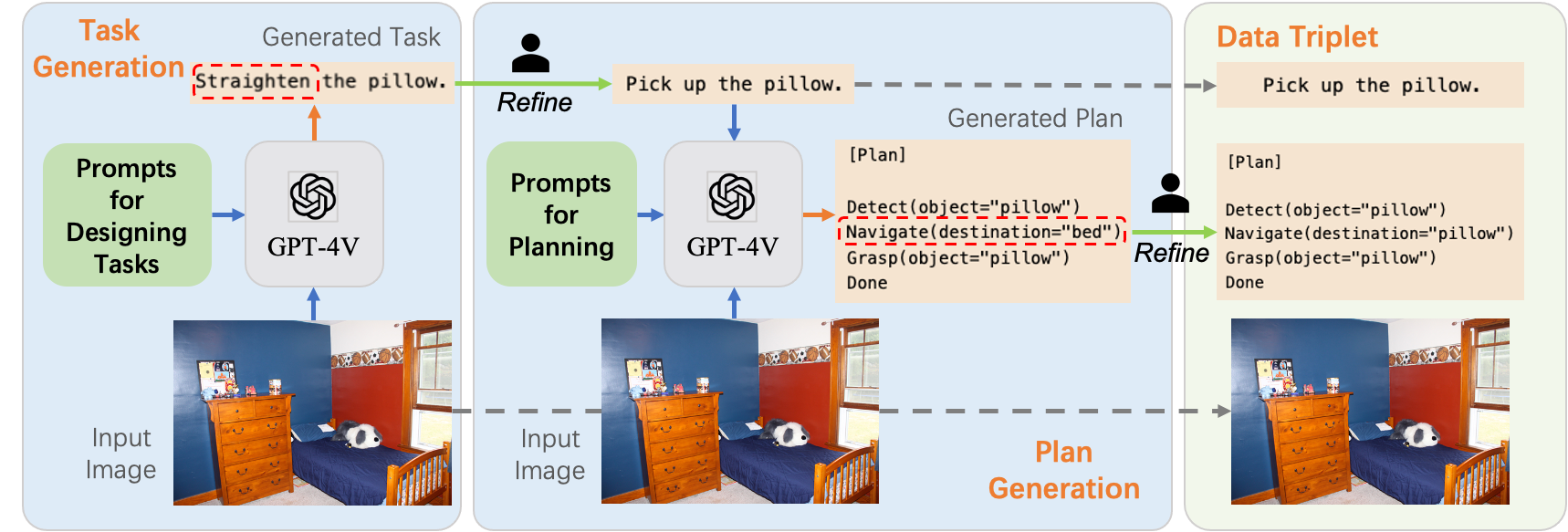} 
\caption{\textbf{Data acquisition pipeline.} We collected images from indoor object detection datasets and used GPT-4V to generate possible tasks a robot may perform on these scenes. We then manually refined the generated tasks and asked GPT-4V to predict corresponding plans. Finally, by manually refining the generated plans, we acquire data triplets of task, plan, and image.}
\label{fig_pipeline}
\end{figure*}

\begin{figure*}[ht]
\centering
\includegraphics[width=0.8\textwidth]{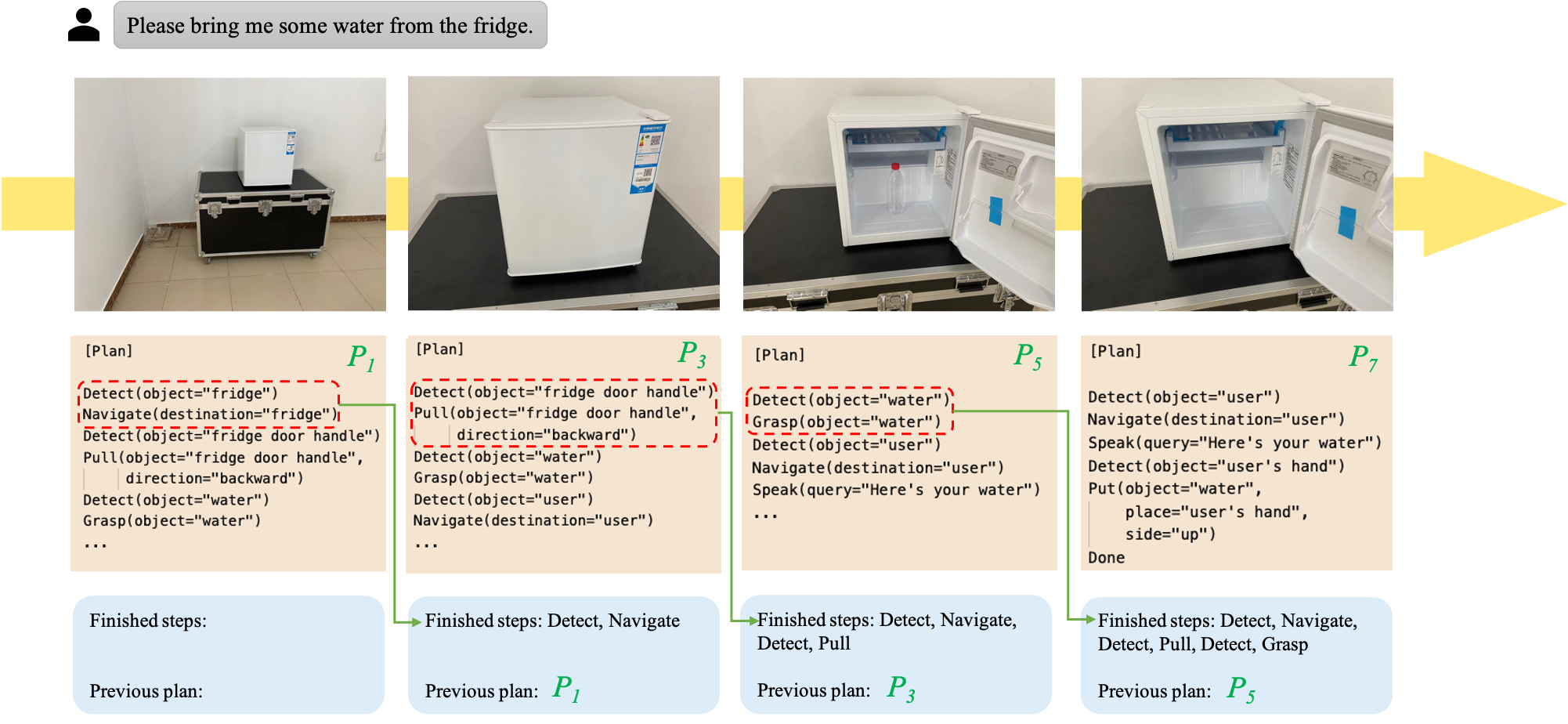} 
\caption{\textbf{Illustration of the planning of ReLEP on the Bring Water task in the real-world experiment using collected images.} Steps that would not change the environment image are omitted during testing.}

\end{figure*}
\begin{table*}[htbp]
    \begin{center}
    \begin{tabular}{l|l}
        \toprule
        \textbf{Skill} & \textbf{Description} \\
        \midrule
        \texttt{Detect(object)} & Detect someone or something referred in the task. \\
        \addlinespace
        \texttt{Grasp(object)} & Grasp the object. \\
        \addlinespace
        \texttt{Navigate(destination)} & Navigate to destination. \\
        \addlinespace
        \texttt{Pull(object, direction)} & Pull object in one of the following directions:up, down, left, right, backward. \\
        \addlinespace
        \texttt{Push(object, direction)} & Push object in one of the following directions:up, down, left, right, forward. \\
        \addlinespace
        \texttt{Put(object, place, side)} & Put the object to the side of the place. \\
        \midrule
        \texttt{Speak(query)} & Communicate with the user by saying the query. \\
        \addlinespace
        \texttt{EQA(query)} & Embodied question answering, return the answer to the query. \\
        \addlinespace
        \texttt{Wait(integer)} & Wait for integer seconds. \\
        \bottomrule
    \end{tabular}
    \caption{\textbf{The Skill Library of ReLEP.} We defined nine skills that can be combined to complete a wide range of tasks in daily life, including interactive skills that can actively change the environment (top part) and non-interactive skills that cannot (bottom part).}
    \label{table:s_l}
    \end{center}
\end{table*}

\begin{figure*}[htbp]
\centering
\includegraphics[width=0.8\textwidth]{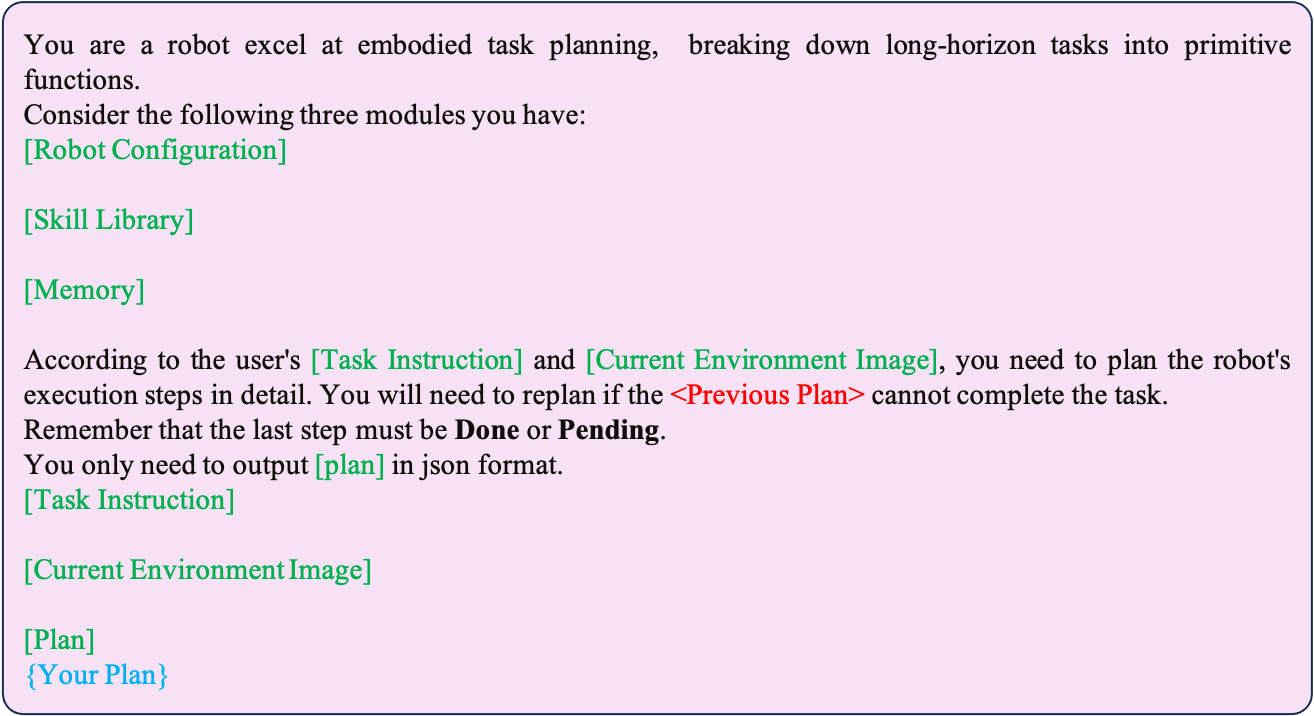}
\caption{\textbf{An outline of ReLEP's system prompt.}}
\end{figure*}

\begin{figure*}[htbp]
\centering
\includegraphics[width=0.8\textwidth]{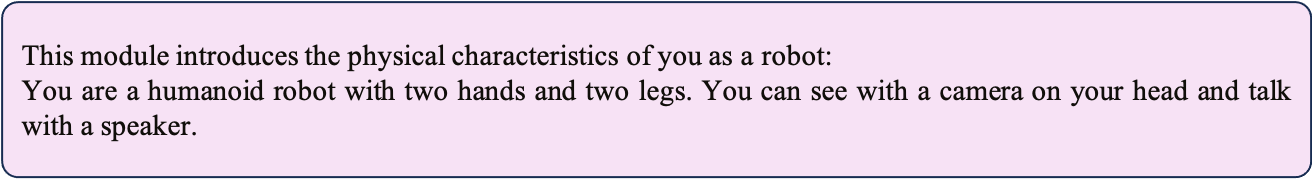}
\caption{\textbf{Prompt of Robot Configuration module.}}
\end{figure*}

\begin{figure*}[htbp]
\centering
\includegraphics[width=0.8\textwidth]{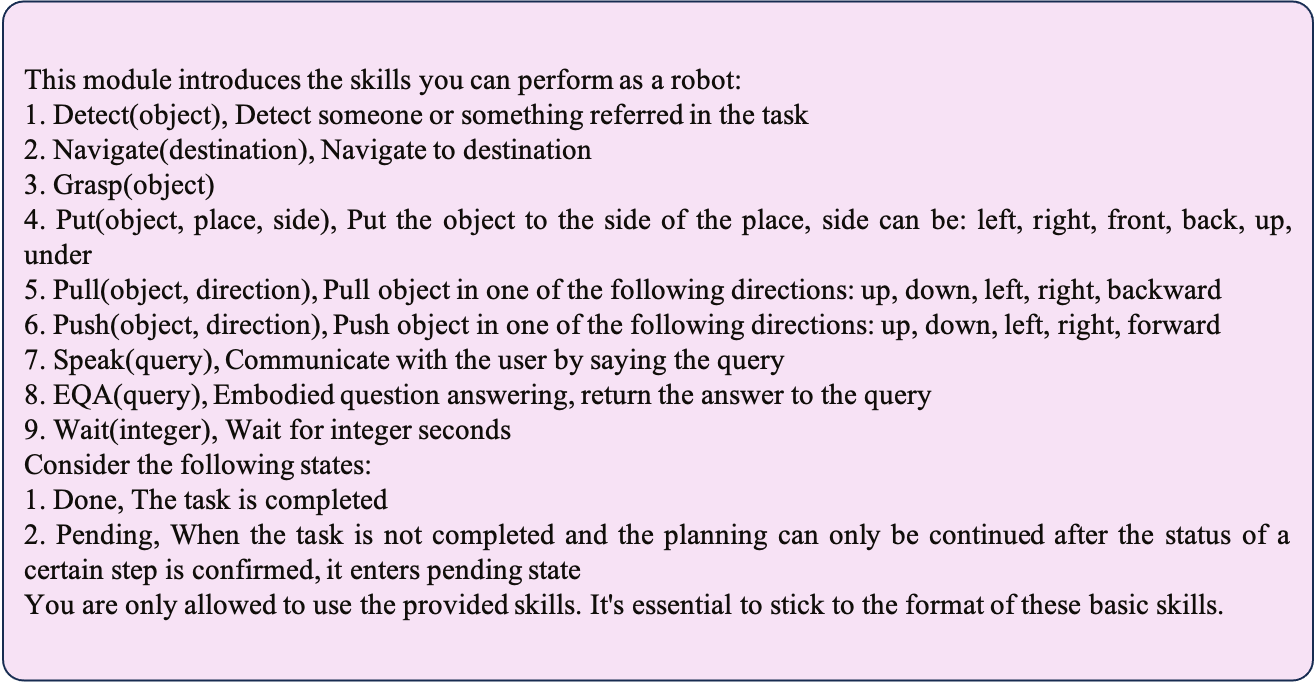}
\caption{\textbf{Prompt of Skill Library module.}}
\end{figure*}

\begin{figure*}[htbp]
\centering
\includegraphics[width=0.8\textwidth]{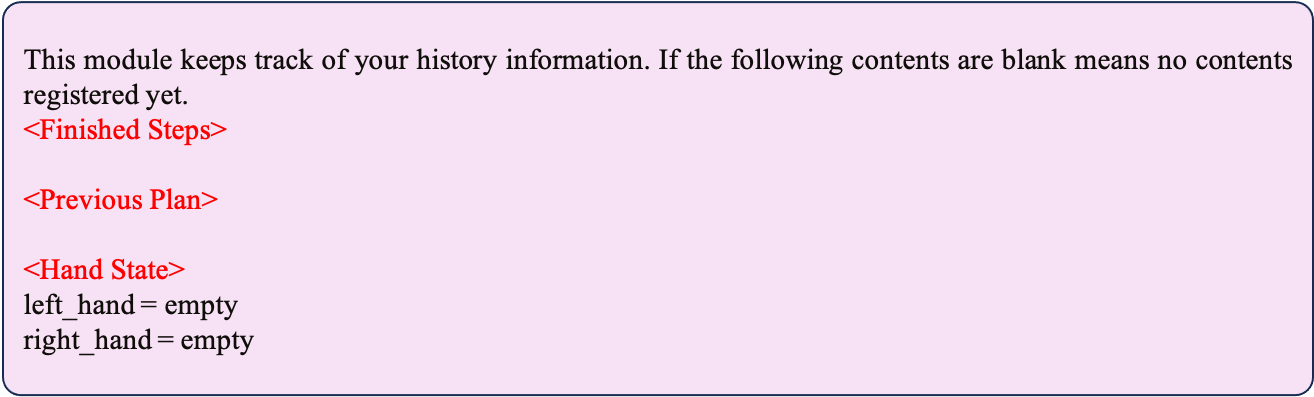}
\caption{\textbf{Prompt of Memory module.}}
\end{figure*}

\begin{table*}[htbp]
    \begin{center}
    \begin{tabular}{l|l}
        \toprule
        \textbf{Task} & \textbf{Description}\\
        \midrule
        Turn on Lights & Turn on the room lights by pressing a light switch. \\
        \addlinespace
        Push Chair & Push an armchair under the table. \\
        \addlinespace
        Bring Water & Bring the user a bottle of water from a closed fridge. \\
        \addlinespace
        Bring Book & Bring the user the book under a cup on the desk. \\
        \addlinespace
        Throw Trash & Pick up the trash on the ground and throw it into the trash can. \\
        \addlinespace
        Fill Kettle & Fill the kettle with tap water. \\
        \addlinespace
        Put Mug & Put the mug into a closed cabinet. \\
        \bottomrule
    \end{tabular}
    \caption{\textbf{Descriptions of the seven long-horizon tasks used in the in-context experiment.}}

    \end{center}
\end{table*}

\begin{figure*}[htbp]
\centering
	\begin{subfigure}[b]{0.49\linewidth}
		\centering
		\includegraphics[width=0.9\linewidth]{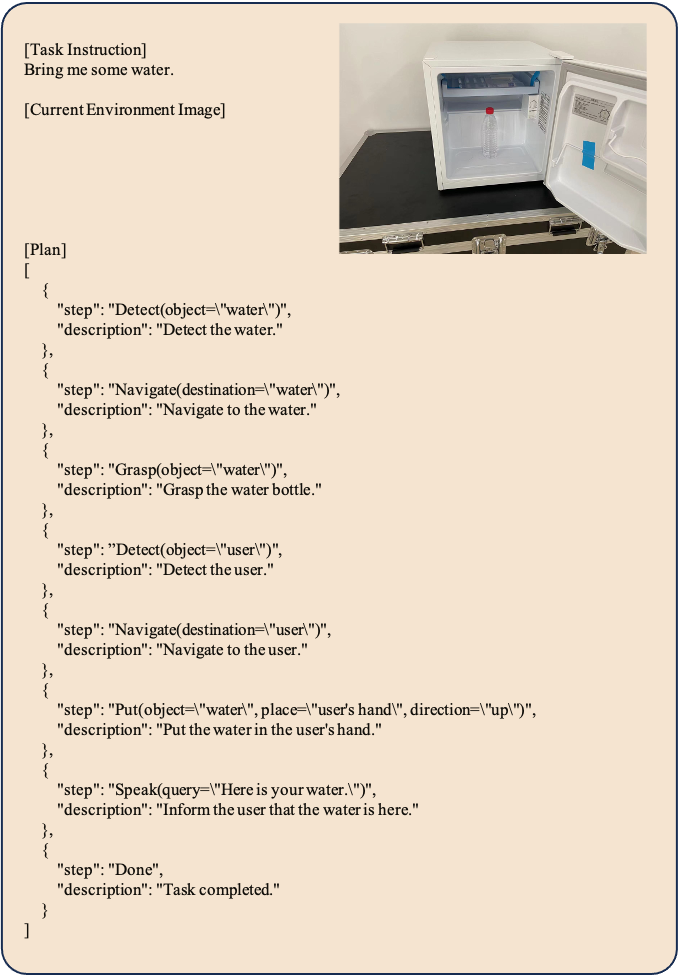}
        \subcaption{A loosely relevant example.}
	\end{subfigure}
    \begin{subfigure}[b]{0.49\linewidth}
		\centering
		\includegraphics[width=0.9\linewidth]{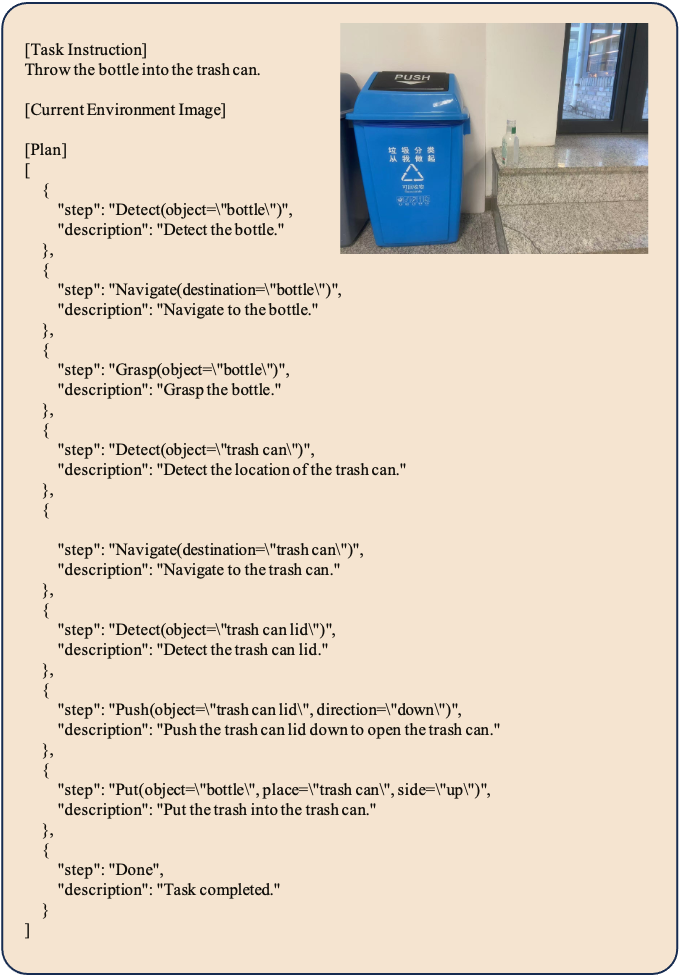}
        \subcaption{A highly relevant example.}
	\end{subfigure}
\caption{\textbf{Illustrations of a loosely relevant example and a highly relevant example of task Throw Trash.} All the loosely relevant examples used in the in-context experiments are the simple Bring Water task. The highly relevant example of task Throw Trash is crafted by replacing the subject trash with bottle.}
\end{figure*}

\begin{table*}[ht]
    \begin{center}
    \begin{tabular}{l|l}
        \toprule
        \textbf{Task} & \textbf{Description}\\
        \midrule
        Close Laptop & Close the laptop with the lid open. \\
        \addlinespace
        Push Chair & Push an armchair under the table. \\
        \addlinespace
        Turn on Lights & Turn on the room lights by pressing a light switch. \\
        \addlinespace
        Look into Mirror & Approach a mirror and tell the user what is in it. \\
        \addlinespace
        Put Mug & Put the mug into a closed cabinet. \\
        \midrule
        Take Egg & Take an egg from a closed fridge and put it on the table. \\
        \addlinespace
        Fill Mug & Fill the mug with tap water and put it on the table. \\
        \addlinespace
        Flush Knife & Flush the knife with tap water. \\
        \addlinespace
        Pull Curtain & Close the shower curtain. \\
        \addlinespace
        Lift Toilet Seat & Lift the toilet seat up. \\
        \bottomrule
    \end{tabular}
    \caption{\textbf{Descriptions of the ten real-time simulated long-horizon tasks.}}

    \end{center}
\end{table*}

\begin{figure*}[ht]
\centering
	\begin{subfigure}[b]{0.24\linewidth}
		\centering
		\includegraphics[width=0.9\linewidth]{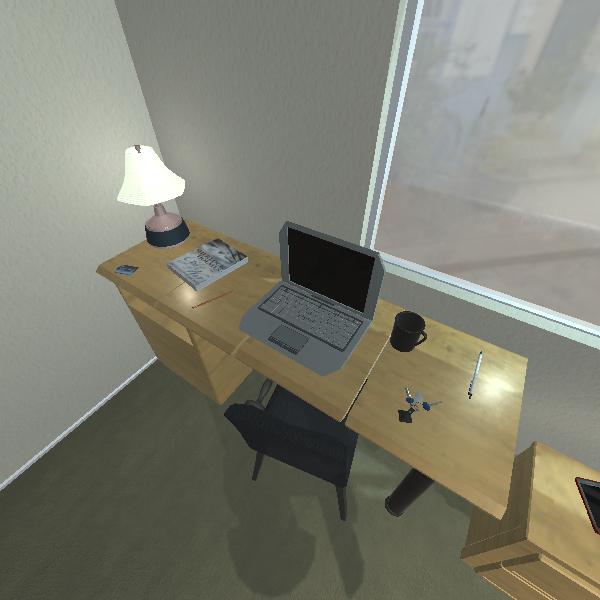}
        \subcaption{Close Laptop}
	\end{subfigure}
    \begin{subfigure}[b]{0.24\linewidth}
		\centering
		\includegraphics[width=0.9\linewidth]{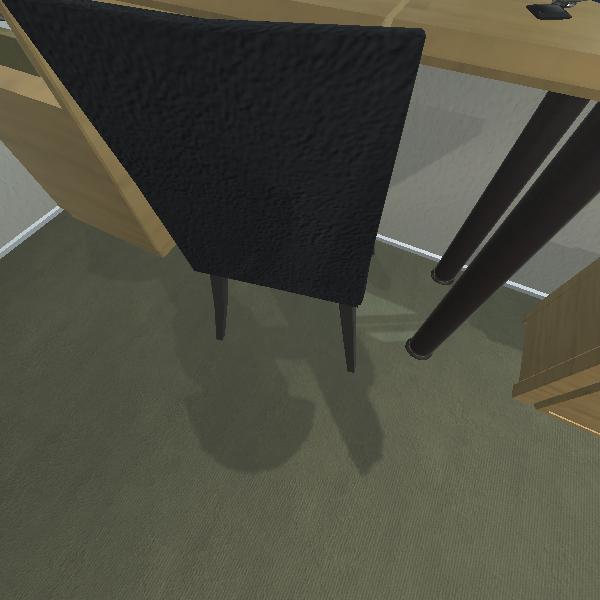}
        \subcaption{Push Chair}
	\end{subfigure}
    \begin{subfigure}[b]{0.24\linewidth}
		\centering
		\includegraphics[width=0.9\linewidth]{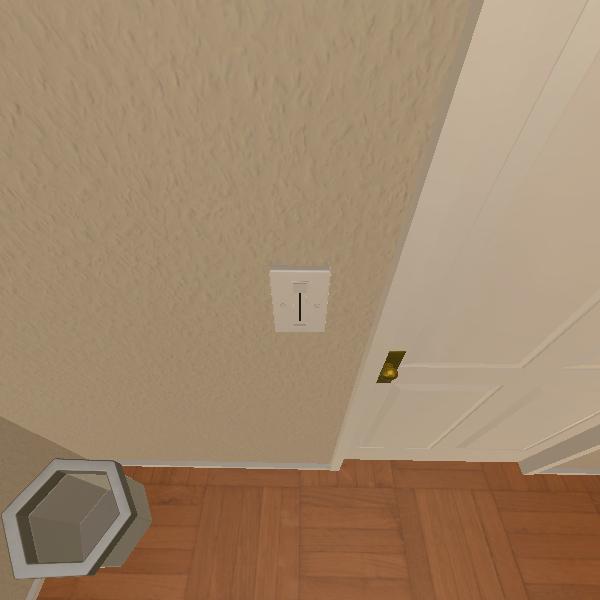}
        \subcaption{Turn on Lights}
	\end{subfigure}
    \begin{subfigure}[b]{0.24\linewidth}
		\centering
		\includegraphics[width=0.9\linewidth]{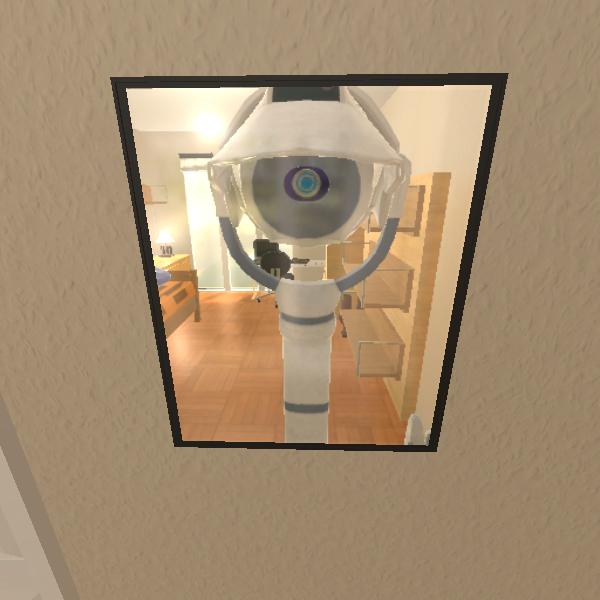}
        \subcaption{Look into Mirror}
	\end{subfigure}
    
    \begin{subfigure}[b]{0.24\linewidth}
		\centering
		\includegraphics[width=0.9\linewidth]{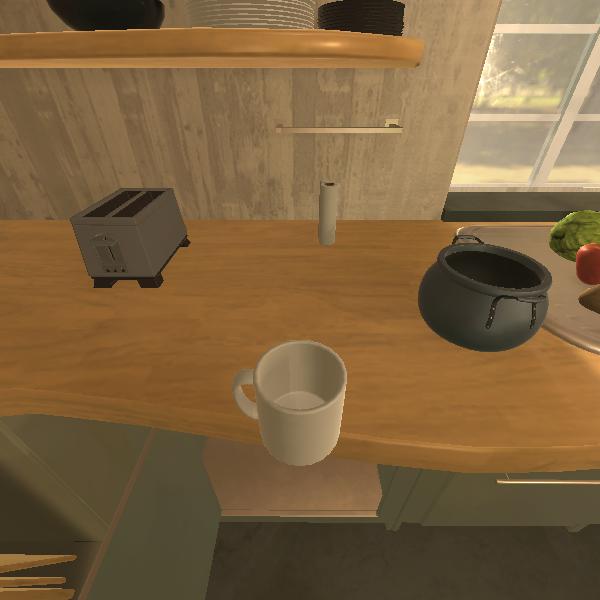}
        \subcaption{Put Mug}
	\end{subfigure}
    \begin{subfigure}[b]{0.24\linewidth}
		\centering
		\includegraphics[width=0.9\linewidth]{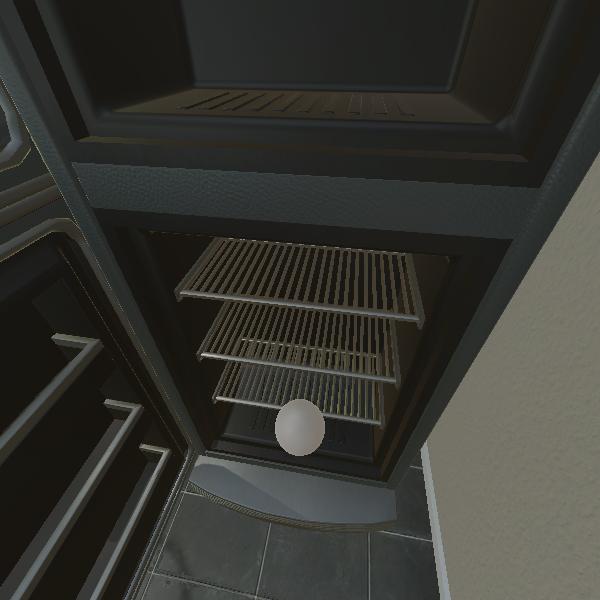}
        \subcaption{Take Egg}
	\end{subfigure}
    \begin{subfigure}[b]{0.24\linewidth}
		\centering
		\includegraphics[width=0.9\linewidth]{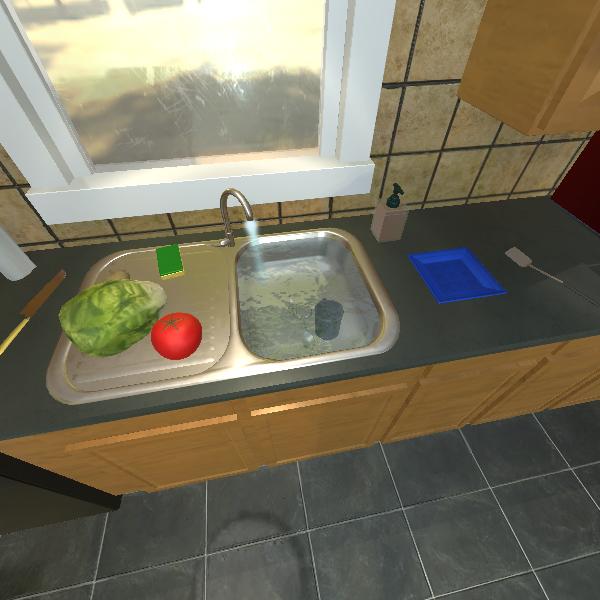}
        \subcaption{Fill Mug}
	\end{subfigure}
    \begin{subfigure}[b]{0.24\linewidth}
		\centering
		\includegraphics[width=0.9\linewidth]{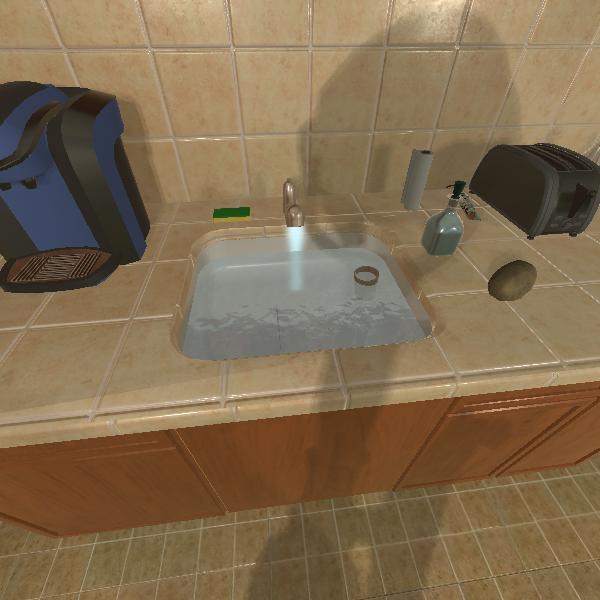}
        \subcaption{Flush Knife}
	\end{subfigure}
    
    \begin{subfigure}[b]{0.24\linewidth}
		\centering
		\includegraphics[width=0.9\linewidth]{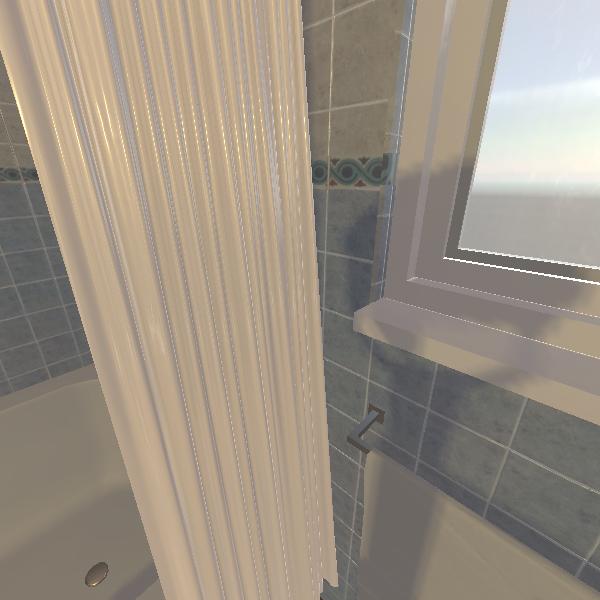}
        \subcaption{Pull Curtain}
	\end{subfigure}
    \begin{subfigure}[b]{0.24\linewidth}
		\centering
		\includegraphics[width=0.9\linewidth]{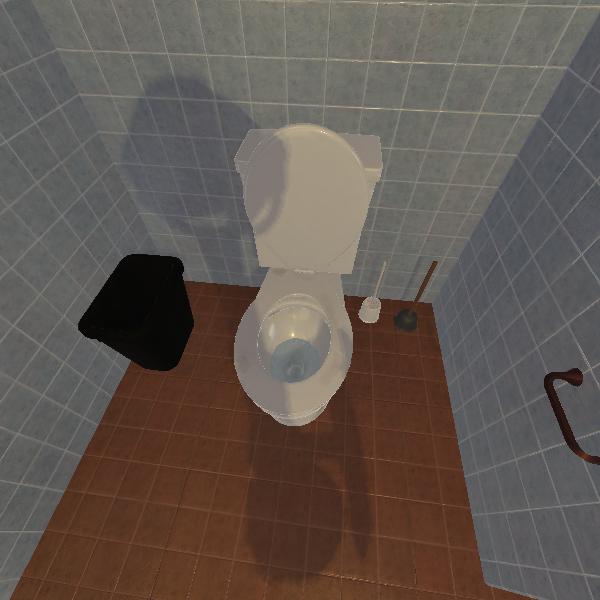}
        \subcaption{Lift Toilet Seat}
	\end{subfigure}
\caption{\textbf{Key frame images of the real-time experiment in simulated environments.}}
\end{figure*}

\begin{table*}[ht]
    \begin{center}
    \begin{tabular}{l|l}
        \toprule
        \textbf{Task Name} & \textbf{Description}\\
        \midrule
        Turn on Lights & Turn on the room lights by pressing a light switch. \\
        \addlinespace
        Push Chair & Push an armchair under the table. \\
        \addlinespace
        Unplug Power & Unplug a power cord from a socket on a vertical wall. \\
        \addlinespace
        Close Laptop & Close the laptop with the lid open. \\
        \addlinespace
        Bring Book & Bring the user a book from the table. \\
        \addlinespace
        Throw Trash & Throw an empty bottle into the trash can. \\
        \addlinespace
        Bring Water & Bring the user a bottled water from a closed fridge. \\
        \addlinespace
        Look into Mirror & Approach a mirror and tell the user what is in it. \\
        \bottomrule
    \end{tabular}
    \caption{\textbf{Descriptions of the eight real-world off-line long-horizon tasks.}}

    \end{center}
\end{table*}

\begin{figure*}[ht]
\centering
	\begin{subfigure}[b]{0.24\linewidth}
		\centering
		\includegraphics[width=0.9\linewidth]{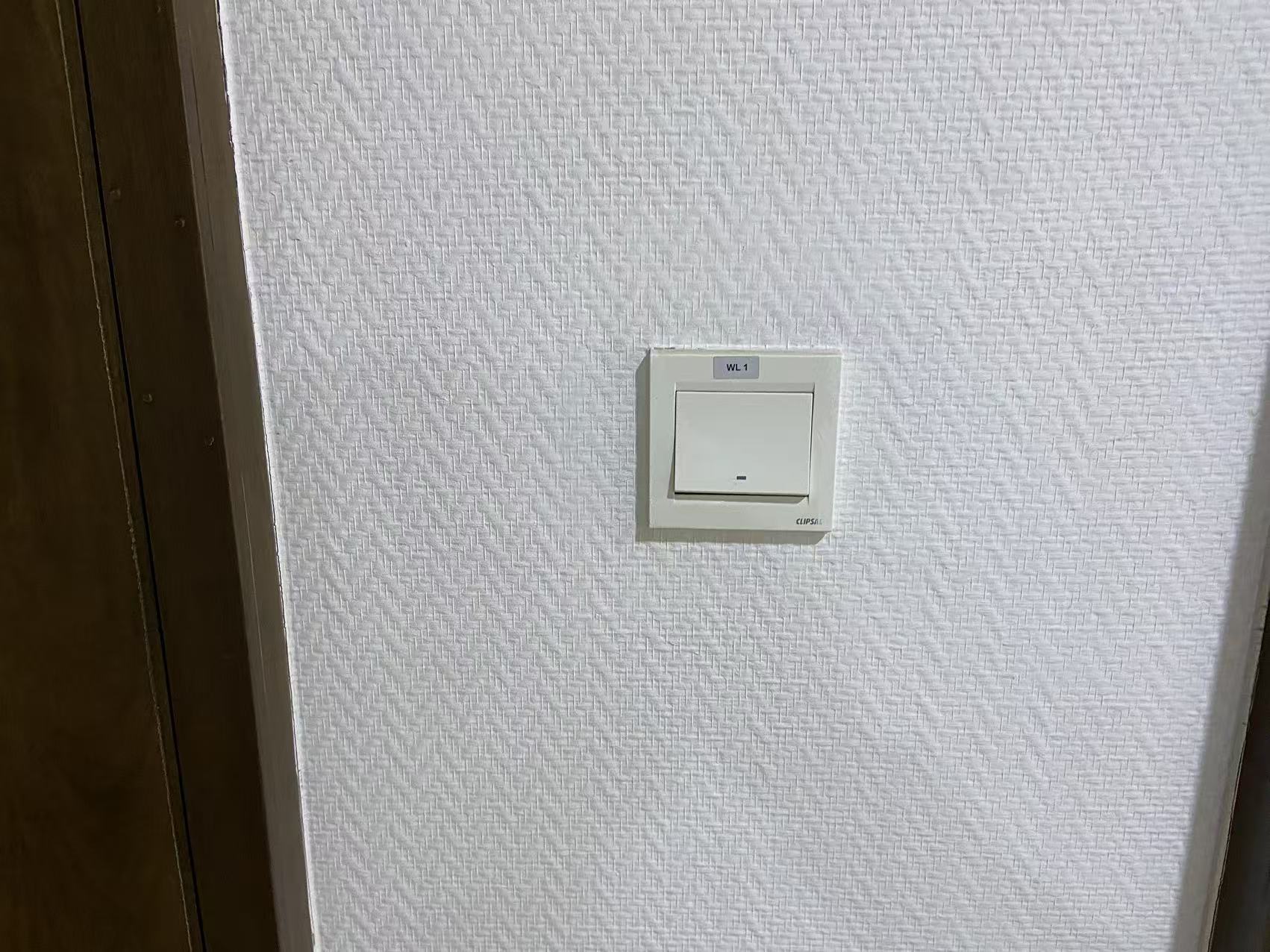}
        \subcaption{Turn on Lights}
	\end{subfigure}
    \begin{subfigure}[b]{0.24\linewidth}
		\centering
		\includegraphics[width=0.9\linewidth]{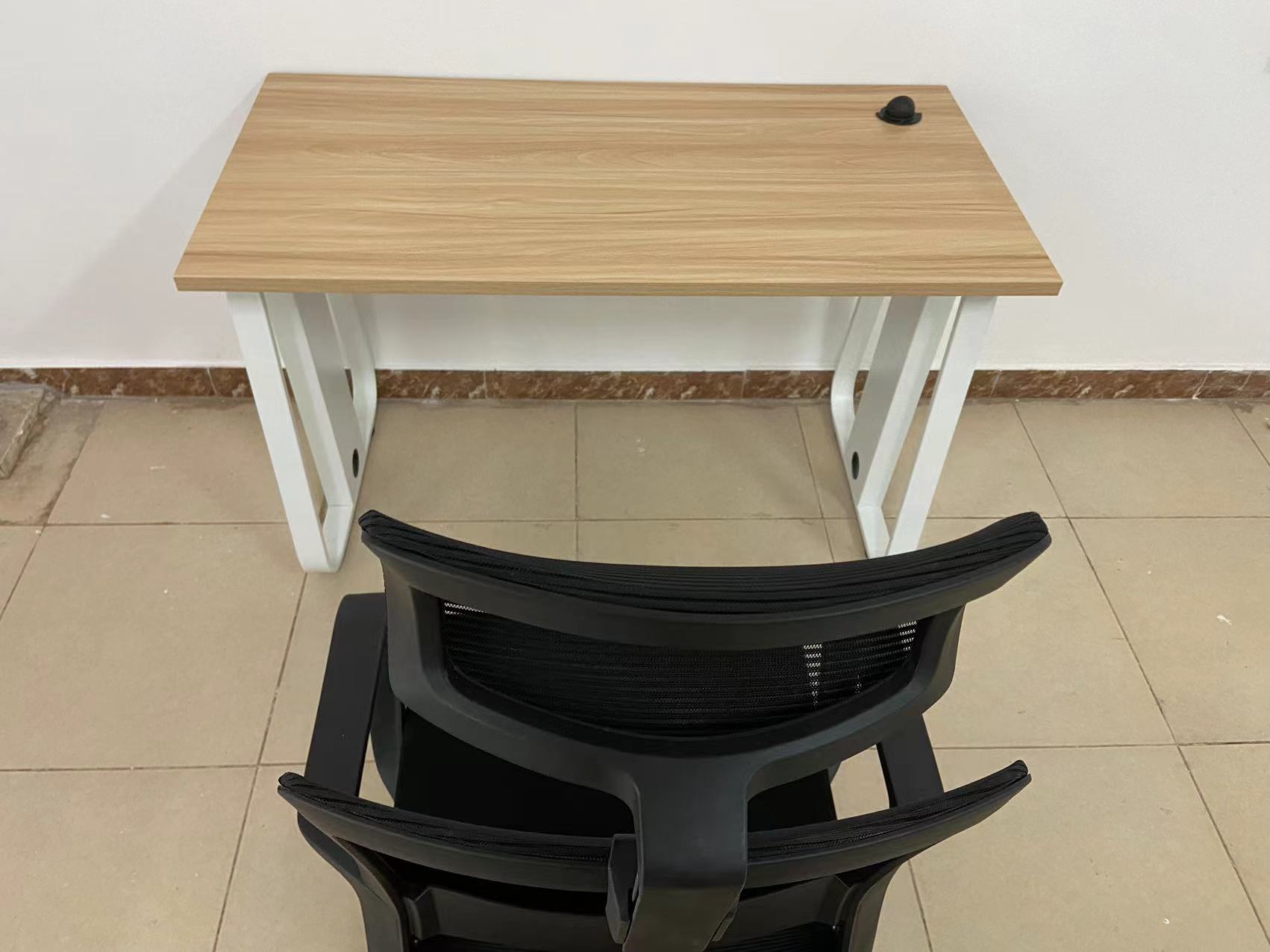}
        \subcaption{Push Chair}
	\end{subfigure}
    \begin{subfigure}[b]{0.24\linewidth}
		\centering
		\includegraphics[width=0.9\linewidth]{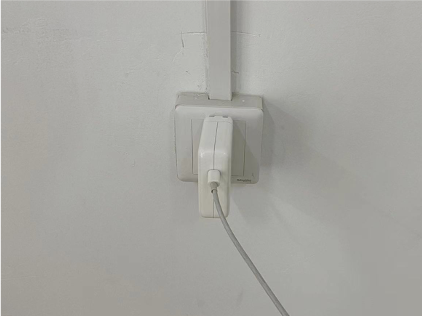}
        \subcaption{Unplug Power}
	\end{subfigure}
    \begin{subfigure}[b]{0.24\linewidth}
		\centering
		\includegraphics[width=0.9\linewidth]{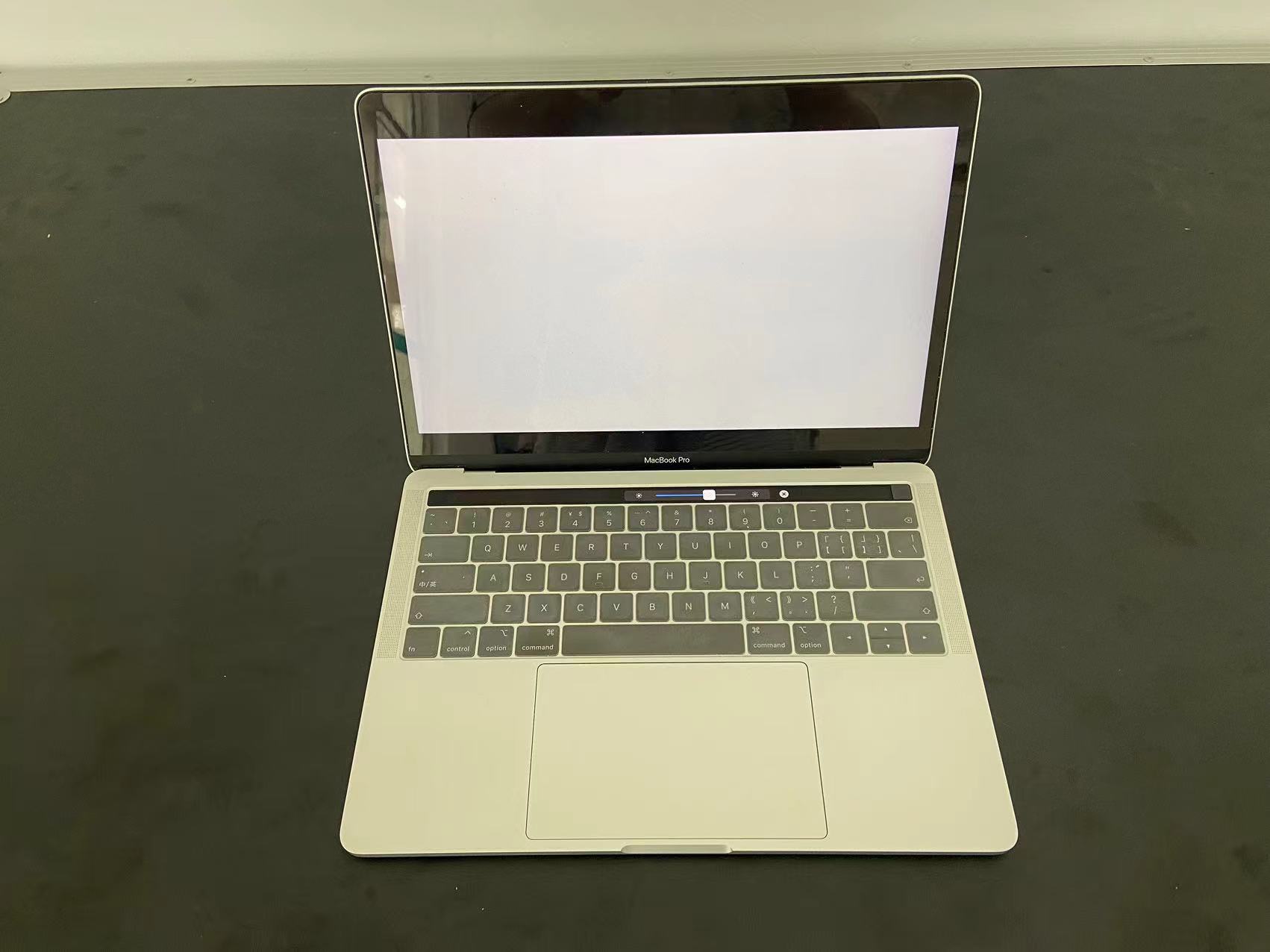}
        \subcaption{Close Laptop}
	\end{subfigure}
    
    \begin{subfigure}[b]{0.24\linewidth}
		\centering
		\includegraphics[width=0.9\linewidth]{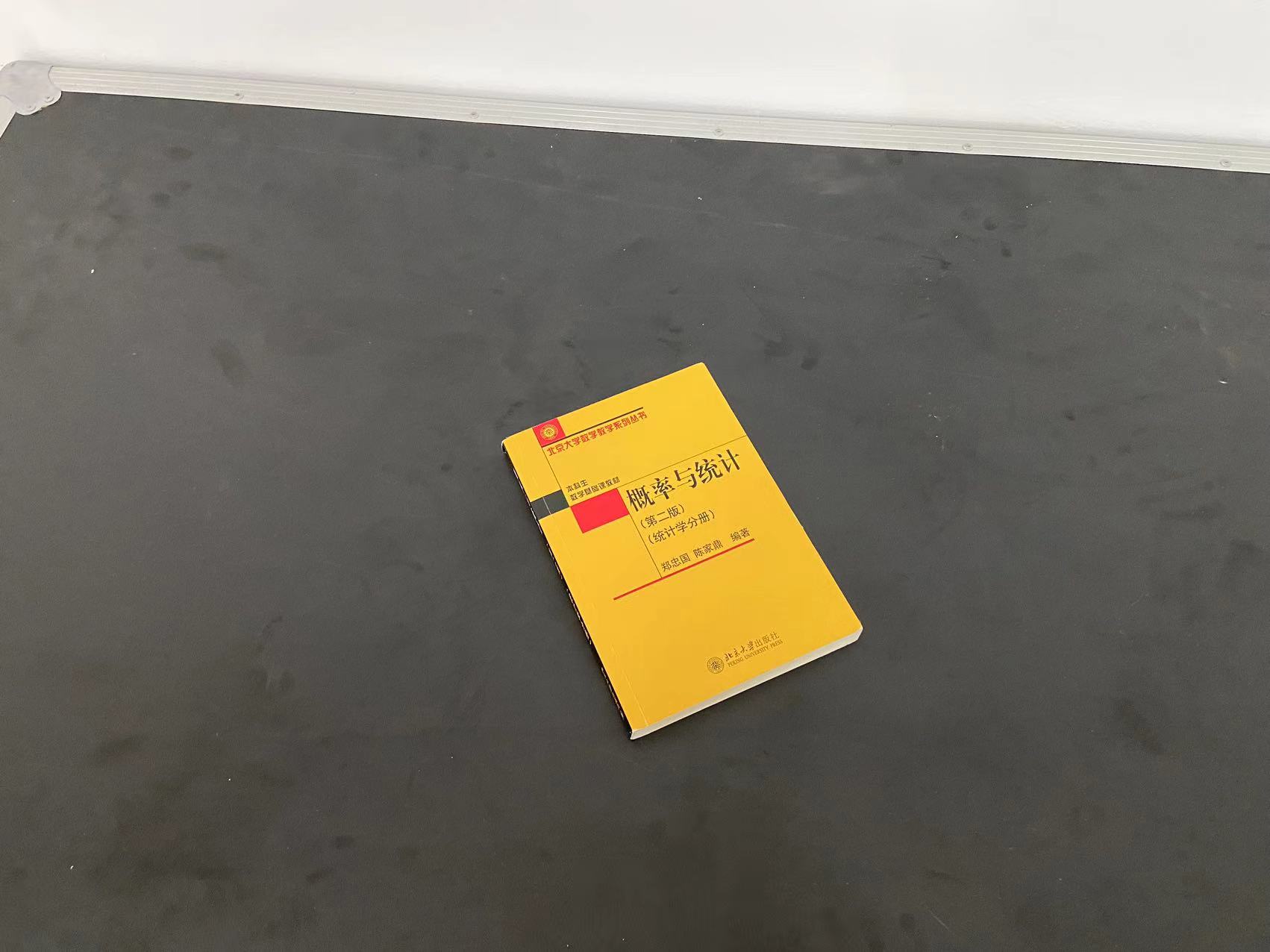}
        \subcaption{Bring Book}
	\end{subfigure}
    \begin{subfigure}[b]{0.24\linewidth}
		\centering
		\includegraphics[width=0.9\linewidth]{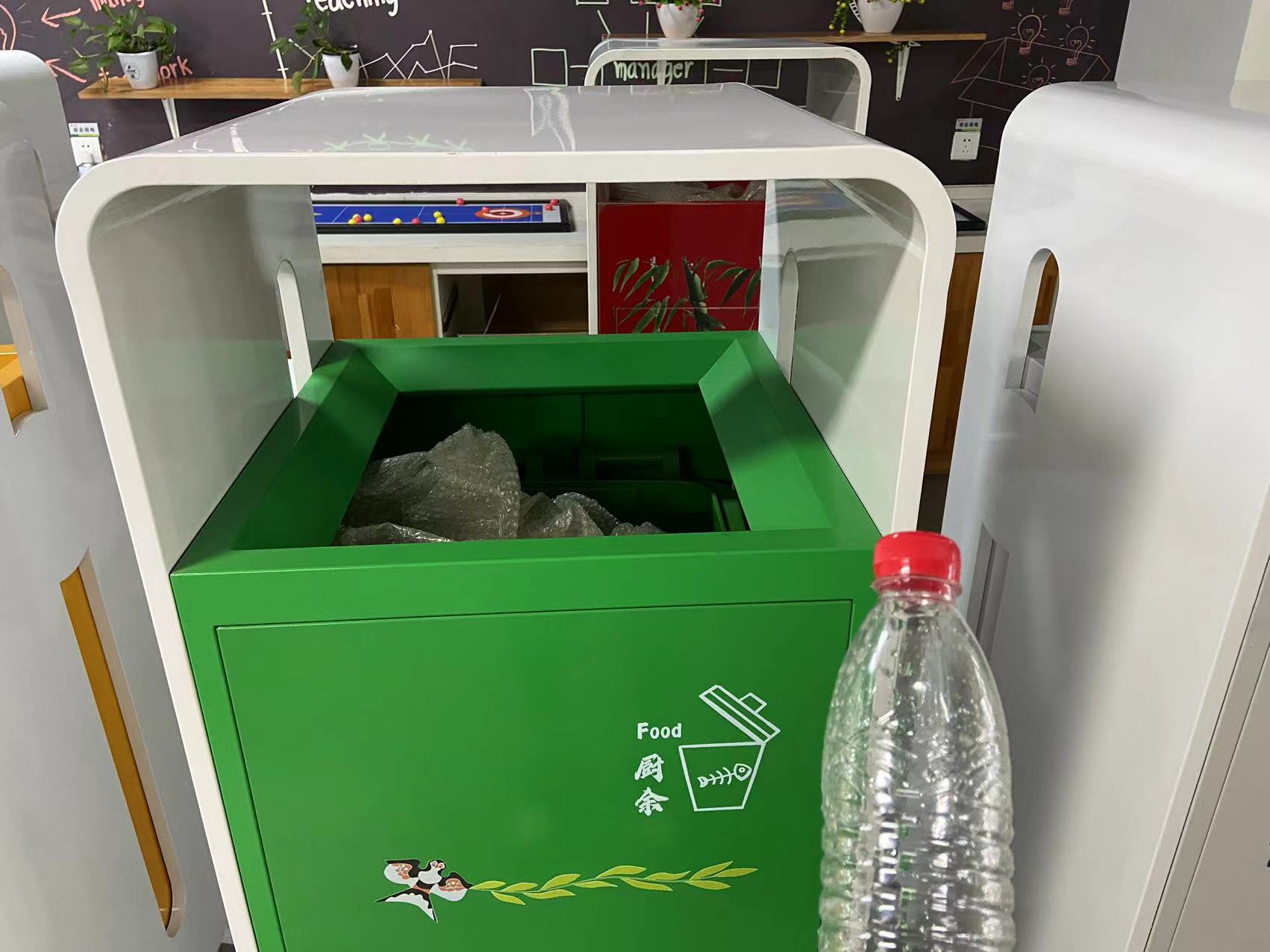}
        \subcaption{Throw Trash}
	\end{subfigure}
    \begin{subfigure}[b]{0.24\linewidth}
		\centering
		\includegraphics[width=0.9\linewidth]{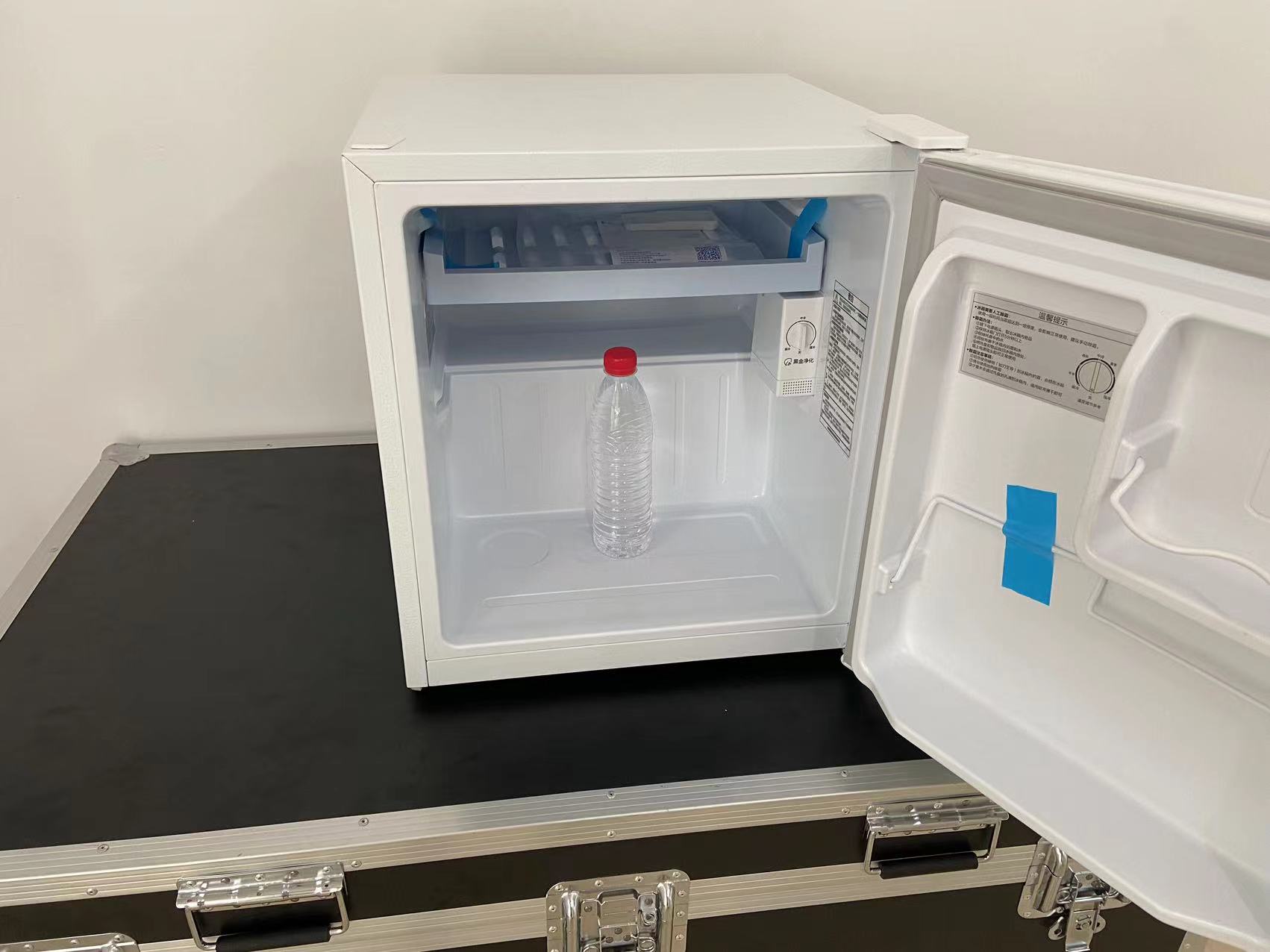}
        \subcaption{Bring Water}
	\end{subfigure}
    \begin{subfigure}[b]{0.24\linewidth}
		\centering
		\includegraphics[width=0.9\linewidth]{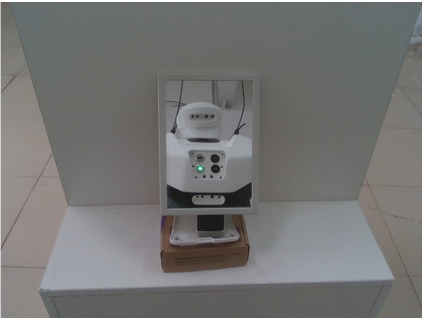}
        \subcaption{Look into Mirror}
	\end{subfigure}
\caption{\textbf{Key frame images of the real-world experiment using collected images.}}
\end{figure*}

\end{document}